\documentclass{article} 
\usepackage{iclr2027_conference,times}

\usepackage{amsmath,amsfonts,bm}

\def\eqref#1{equation~\ref{#1}}

\def\1{\bm{1}}

\DeclareMathAlphabet{\mathsfit}{\encodingdefault}{\sfdefault}{m}{sl}
\SetMathAlphabet{\mathsfit}{bold}{\encodingdefault}{\sfdefault}{bx}{n}

\usepackage{hyperref}
\usepackage{url}
\usepackage{graphicx}
\usepackage{subcaption}
\usepackage{multirow}
\usepackage{algorithm}
\usepackage{algpseudocode}
\usepackage{amssymb}

\title{Quantization-Aware Pre-Training with Constrained Empirical Weight Distribution}

\author{Ningfeng Yang \\
University of British Columbia\\
\texttt{nxyang@ece.ubc.ca} \\
\And
Tor M. Aamodt \\
University of British Columbia\\
\texttt{aamodt@ece.ubc.ca}
}

\iclrfinalcopy 
\begin{document}

\maketitle

\begin{abstract}
Quantization-Aware Pre-Training (QAPT) can increase the inference efficiency of DNNs, but a problematic behaviour known as rounding boundary weight oscillation can introduce detrimental noise into the training process and significantly reduce convergence speed. While existing methods can reduce this detrimental noise, they either introduce additional hyperparameters or memory overhead, or cannot consistently improve model accuracy. In this work, we propose optimization with \textbf{C}onstrained \textbf{E}mpirical \textbf{W}eight dis\textbf{T}ribution (CEWT), the first hyperparameter-free memory-overhead-free oscillation suppression method that consistently improves QAPT performance: an optimizer post-update step that projects weights to the nearest point in weight space whose empirical distribution (histogram) matches a zero-mean Gaussian. Our key insight is many quantizers are designed with the implicit assumption that the to-be-quantized data are permutations of samples from a zero-mean Gaussian, and this assumption is not true during QAPT. By enforcing the zero-mean Gaussian prior as a hard constraint, CEWT can suppress this detrimental noise. Empirical results on various combinations of SOTA quantizers and hypersphere optimizers suggest, that with a geomean increase of 4\% in training time, CEWT can consistently reduce the pre-training perplexity (by an average of 2.5 and up to 21 points) of low-precision (down to 1-bit activations and weights and up to 610M parameters) LLaMA/GPT models  without introducing any hyperparameters or storage overhead. Code is available at \url{https://github.com/1733116199/cewt}.
\end{abstract}

\begin{figure}[b]
    \centering
    \begin{subfigure}[t]{0.68\textwidth}
        \includegraphics[width=\textwidth]{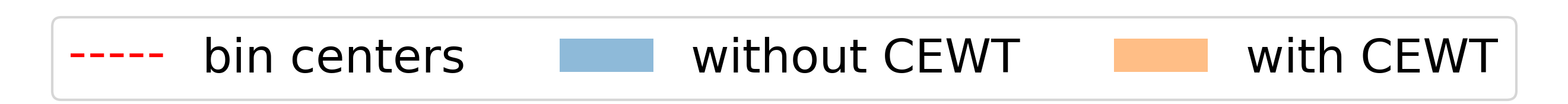}
    \end{subfigure}
    \begin{subfigure}[t]{0.328\textwidth}
        \includegraphics[width=\linewidth]{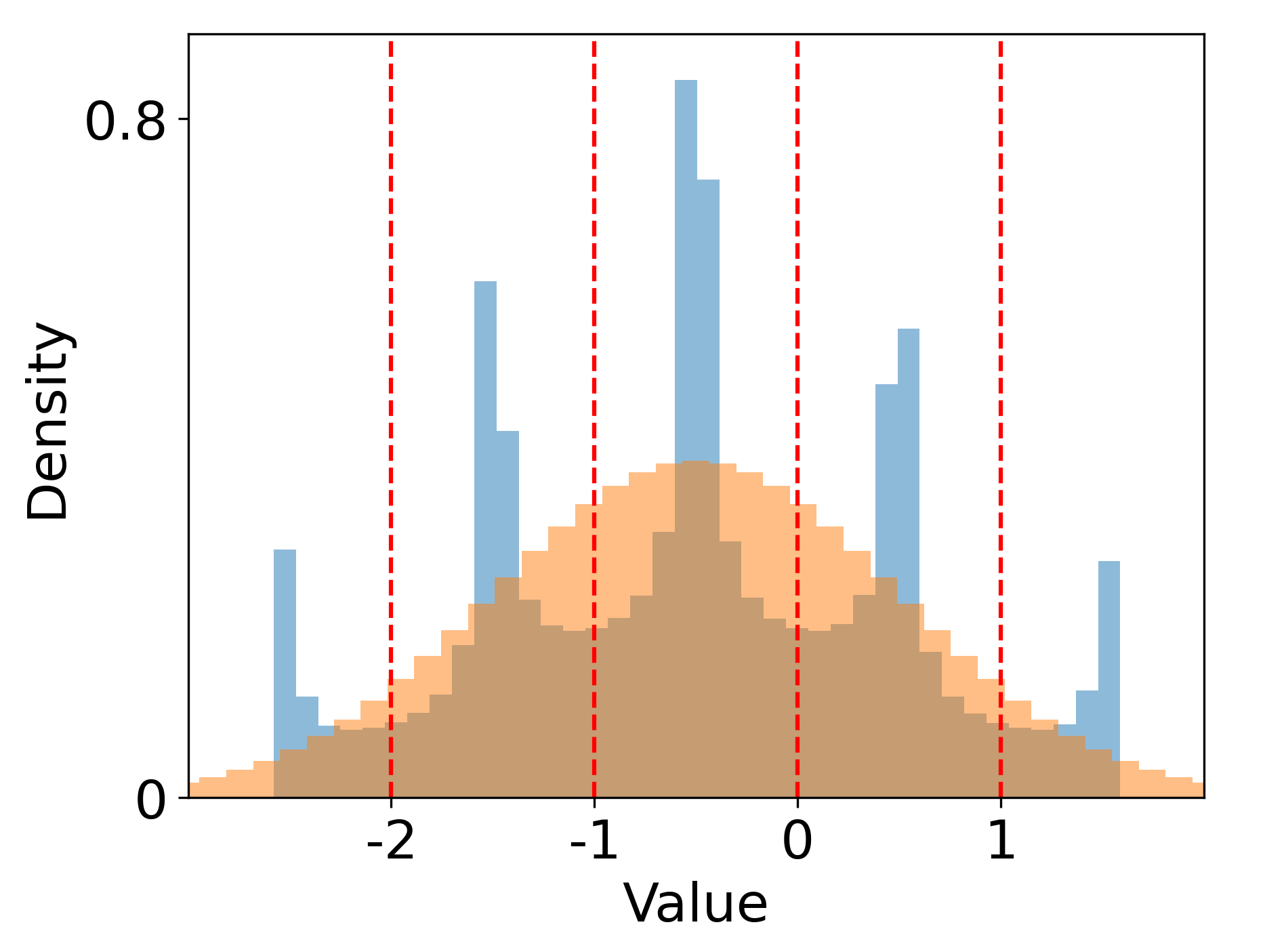}
        \caption{QuEST~\citep{panferov2025queststabletrainingllms}}
        \label{fig:intro_a}
    \end{subfigure}
    \begin{subfigure}[t]{0.3\textwidth}
        \includegraphics[width=\linewidth]{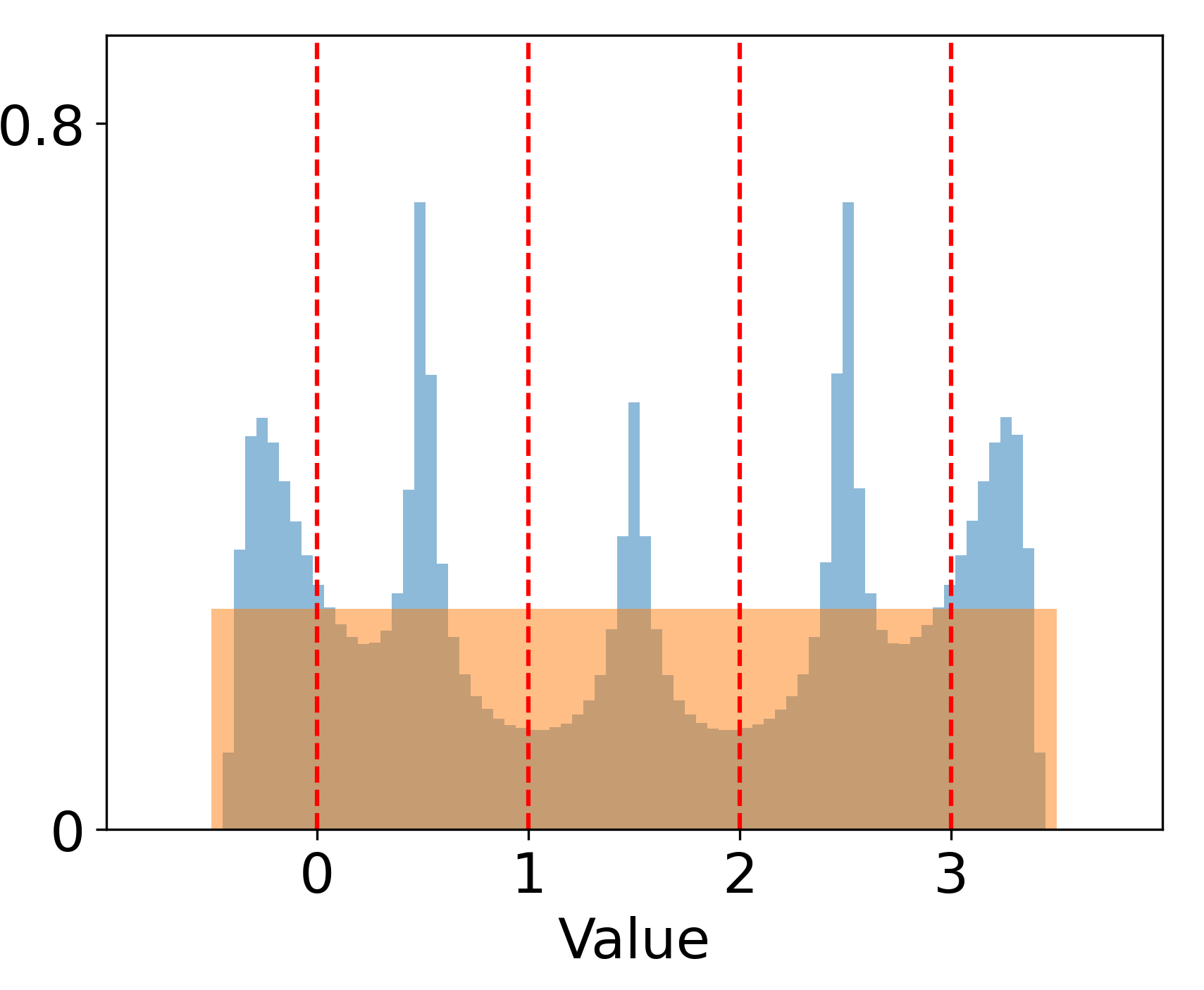}
        \caption{BBQ~\citep{yang2026boosting}}
        \label{fig:intro_b}
    \end{subfigure}
    \begin{subfigure}[t]{0.3\textwidth}
        \includegraphics[width=\linewidth]{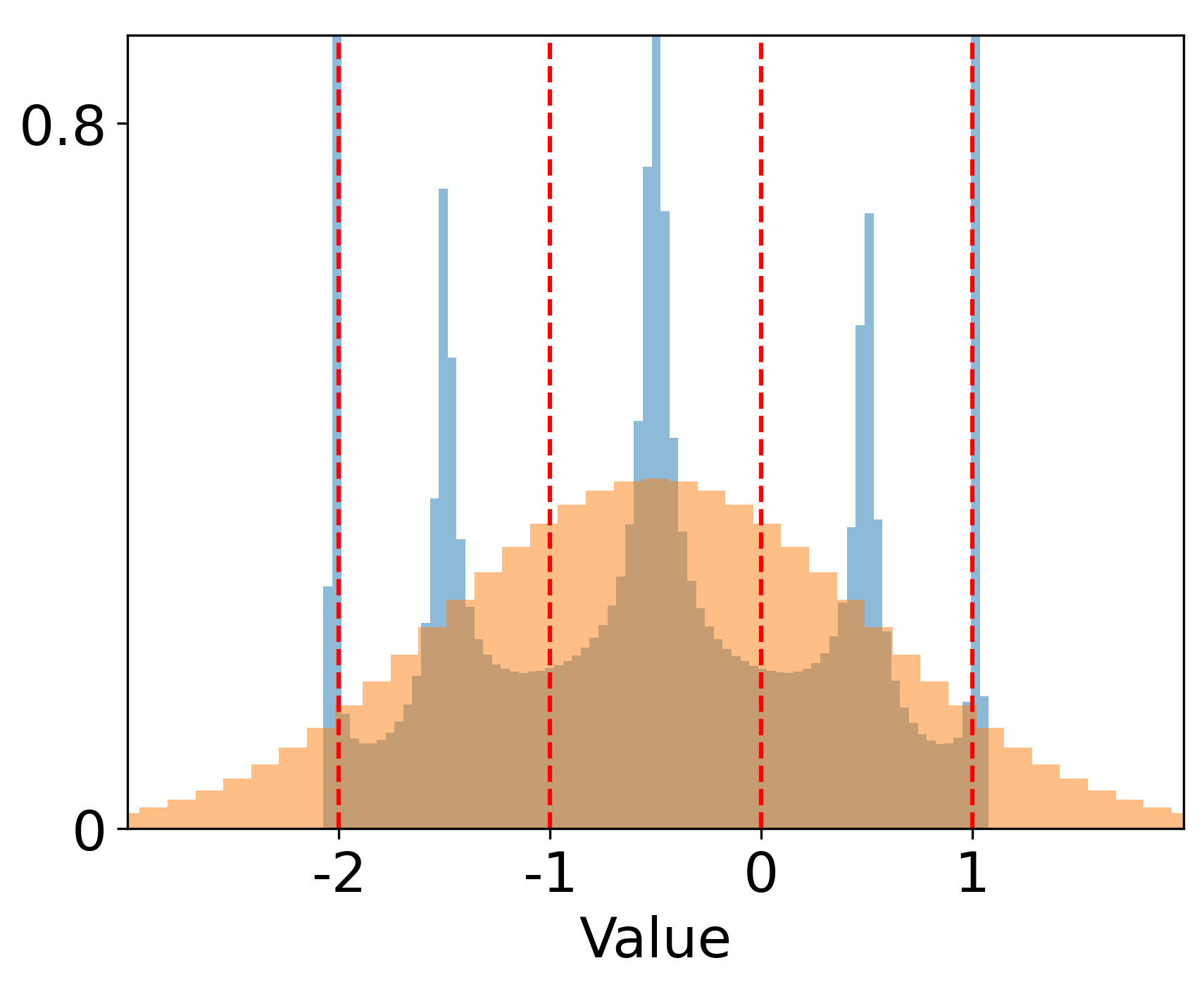}
        \caption{LSQ~\citep{esser2020learnedstepsizequantization}}
        \label{fig:intro_c}
    \end{subfigure}
    \caption{Effects of CEWT on pre-round (BBQ) and pre-clip (QuEST/LSQ) weight distributions of a LLaMA-95M with 2-bit weights and activations pre-trained on 3 billion C4 tokens. See Algorithms~\ref{alg:bbq}, \ref{alg:quest}, and \ref{alg:lsq} from definitions of pre-round and pre-clip weights.
    }
    \label{fig:intro}
\end{figure}
\section{Introduction}
Quantization can reduce the inference complexity of Deep Neural Networks (DNNs). Compared to full-precision DNNs, quantized DNNs consume lower memory capacity and communication bandwidth in data centers, while simultaneously achieving higher inference throughput and energy efficiency when deployed to edge devices. However, naively quantizing DNNs often degrades their quality. While Post-Training Quantization (PTQ) methods~\citep{liu2025spinquant,lin2024duquantdistributingoutliersdual,shao2024omniquantomnidirectionallycalibratedquantization,ma2024affinequantaffinetransformationquantization,liu2024qllmaccurateefficientlowbitwidth,kim2024squeezellmdenseandsparsequantization} can mitigate quality degradation at higher precisions without re-training, PTQ methods generally result in a large accuracy gap when both activations and weights are quantized to 4-bit and below~\citep{panferov2025queststabletrainingllms,kumar2024scalinglawsprecision,ma2024era1bitllmslarge,esser2020learnedstepsizequantization}. In contrast, Quantization-Aware Training (QAT) methods~\citep{liu2025paretoq,panferov2025queststabletrainingllms,ma2024era1bitllmslarge,kumar2024scalinglawsprecision,DBLP:journals/corr/abs-2402-10787,9577791,esser2020learnedstepsizequantization} can narrow this accuracy gap~\citep{du-etal-2024-bitdistiller,panferov2025queststabletrainingllms,liu-etal-2024-llm} by introducing quantization into the optimization process. QAT can be further categorized into Quantization-Aware Pre-Training~\citep{panferov2025queststabletrainingllms,yang2025improving} (QAPT) and Quantization-Aware Fine-Tuning~\citep{malinovskii2024pvtuning,du-etal-2024-bitdistiller} (QAFT). 
QAFT starts from a full-precision pre-trained checkpoint and briefly trains a low-precision model on a comparatively small~\citep{du-etal-2024-bitdistiller} downstream dataset.
In contrast, QAPT trains a low-precision model from scratch on a much larger dataset and for substantially longer.
Compared with full-precision pre-training followed by PTQ or QAFT, QAPT may accelerate pre-training~\citep{kumar2024scalinglawsprecision,xi2023training,DBLP:journals/corr/abs-2505-14669} by performing the forward pass in low precision.
This work focuses on improving the accuracy of QAPT when both weights and activations are quantized to 2-bits and below.

In QAT, a problem known as rounding boundary weight oscillation~\citep{pmlr-v162-nagel22a,10.5555/3618408.3619311,wenshj2025oscillations}, often attributed~\citep{pmlr-v162-nagel22a} to the gradient bias of the Straight-Through Estimator~\citep{bengio2013estimatingpropagatinggradientsstochastic} and other rounding Jacobian estimators, can introduce detrimental noise into the optimization process and slow down convergence, especially at the later stages of training. A significant portion of parameters gather at the rounding boundaries of the quantization grid, and constantly oscillate from one side of the rounding boundaries to the other. 
Existing methods attempt to address the slow convergence problem by freezing oscillating parameters~\citep{pmlr-v162-nagel22a}, by reducing the number of parameters sufficiently close to rounding boundaries~\citep{pmlr-v162-nagel22a,10.5555/3618408.3619311,li2026winqacceleratingquantizationawaretraining}, or by simulating deterministic quantization error with random noise~\citep{Shin_2023_CVPR}. However, these methods either introduce additional hyperparameters, which when under-tuned can reduce the effectiveness, while proper tuning can increase training budget, or they cannot consistently improve the performance of QAPT due to the use of biased optimization objectives.

Inspired by Hypersphere Optimization~\citep{hyperballOptimization2026,xie2026controlled,ren2026rethinkinglanguagemodelscaling}, which removes weight decay as a hyperparameter by enforcing what weight decay \textit{does} (i.e. controlling the Frobenius norm of weights) as a hard constraint, we propose optimization with \textbf{C}onstrained \textbf{E}mpirical \textbf{W}eight Dis\textbf{T}ribution (CEWT, pronounced \textit{cute}), a hyperparameter-free extension to existing Hypersphere optimizers (e.g. AdamH~\citep{kingma2017adammethodstochasticoptimization,hyperballOptimization2026}, MuonH~\citep{muon,hyperballOptimization2026}, LionH~\citep{chen2023symbolicdiscoveryoptimizationalgorithms,hyperballOptimization2026}, and SSO~\citep{xie2026controlled}) that projects weights to the nearest point in weight space whose empirical distribution (histogram) matches a Gaussian. CEWT is inspired by our observation that many quantizers~\citep{yang2026boosting,panferov2025queststabletrainingllms,dettmers2023qloraefficientfinetuningquantized,10.1007/978-3-031-20083-0_23} are designed based on the assumption that the empirical distribution of raw weights are, or can be approximately converted to, a bell-shaped distribution approximately centered around zero --- an assumption that is \textit{not} enforced, and is generally \textit{false}, in practice. 
To address this gap between assumption and reality, CEWT enforces the distribution shape to match a zero-mean Gaussian as a hard constraint. CEWT can suppress rounding boundary weight oscillation as it regulates, \textit{by design}, the number of parameters near rounding boundaries and can significantly accelerate convergence without introducing any hyperparameters.

\section{Background and Motivation}
In this section, we discuss the rounding boundary weight oscillation problem and existing solutions, and Hypersphere Optimization which removes the weight decay hyperparameter by adding a hard constraint to the optimization problem. As we will discuss later in Section~\ref{sec:proposed}, CEWT also removes a hyperparameter by adding a hard constraint. 

\subsection{Rounding boundary Weight Oscillation}\label{sec:oscillation}
Rounding boundary weight oscillation~\citep{pmlr-v162-nagel22a} is the phenomenon that during QAT, a significant portion of parameters gather at a rounding boundary, and constantly move from one side of the rounding boundary to the other, generating significant noise which prevents convergence. Weight oscillation manifests as irregularly high density spikes near rounding boundaries~\citep{pmlr-v162-nagel22a,10.5555/3618408.3619311}, and can be detected by plotting the histogram of pre-rounding weights.
Unless explicitly addressed, weight oscillation happens regardless of the quantizer used. For example, Figure~\ref{fig:intro} shows the pre-rounding weight distribution of three SOTA quantizers and their density spikes near rounding boundaries. Weight oscillation tends to happen when a parameter is close to a locally optimal value~\citep{pmlr-v162-nagel22a}. Weight oscillation happens regardless~\citep{pmlr-v162-nagel22a} of the rounding Jacobian estimation method\footnote{Section~\ref{sec:toy_grad_est} and Table~\ref{tab:toy_grad_est} empirically shows that some rounding Jacobian estimators may be less vulnerable, although not immune, to rounding boundary weight oscillation, and that CEWT can suppress rounding boundary weight oscillation and improve accuracy \textbf{regardless} of the rounding Jacobian estimator used. }. Section~\ref{sec:eval_deep} empirically demonstrates rounding boundary weight oscillation cannot be effectively suppressed with the use of optimizers equipped with momentum (e.g. MuonH/LionH) and/or preconditioning (e.g. AdamH/MuonH). In DNN pre-training, where parameters are typically interdependent\footnote{Sections~\ref{sec:exp_2d} and \ref{sec:toy_expl} discuss parameter interdependency and its impact on QAT.}, a subset of oscillating parameters can introduce noise into the gradient of other non-oscillating parameters. The gradient noise becomes more severe as precision (number of bits) reduces and the distance between two quantization bins increases.  Section~\ref{sec:oscillation_example} shows two simple example problems that demonstrate the root cause of weight oscillation.

Existing methods are effective in oscillation suppression but may degrade model accuracy and introduce additional hyperparameters and memory overhead. \citet{pmlr-v162-nagel22a} proposes to detect oscillating weights and freeze them until the end of training. Weight freezing requires additional storage to record the history of weights for oscillation detection and an oscillating frequency threshold hyperparameter which when under-tuned, as \citet{pmlr-v162-nagel22a} noted, can freeze too many parameters too early, degrading the model's final accuracy. \citet{pmlr-v162-nagel22a} also proposed an oscillation dampening loss to regularize the squared distance between quantized weights and clipped weights, effectively pulling unclipped parameters away from rounding boundaries, and can therefore prevent oscillation. However, \citet{pmlr-v162-nagel22a} noted that overly strong regularization can inhibit beneficial weight movement between quantization bins and harm final accuracy. \citet{10.5555/3618408.3619311} proposes Confidence-Guided Annealing (CGA), which trains only parameters sufficiently close to rounding boundaries while freezing parameters far from boundaries.
CGA requires the schedule of a hyperparameter that defines ``sufficiently close" to rounding boundaries. \citet{10.5555/3618408.3619311} stated that when the threshold hyperparameter is overly small, too many parameter are frozen, inhibiting beneficial weight movement which leads to lower final accuracy. 
\citet{li2026winqacceleratingquantizationawaretraining} proposed to periodically re-initialize weights to a linear interpolation between weights and quantized weights and showed this interpolation is equivalent to regularizing the squared distance between quantized and raw weights. While not originally proposed to suppress weight oscillation, empirically we show Periodic Interpolation also suppresses oscillation (Section~\ref{sec:eval_toy}) and we hypothesize this is due to the theoretical connection to dampening loss---both periodic interpolation and dampening loss explicitly pulls weights away from rounding boundaries.
Periodic interpolation requires an interpolation ratio hyperparameter. \citet{li2026winqacceleratingquantizationawaretraining} noted that too large or too low interpolation ratio can lead to suboptimal performance. Pseudo-Quantization Noise (PQN)~\citep{Shin_2023_CVPR} simulates the behaviour of rounding with addition of noise from $U(-0.5, 0.5)$, effectively replacing the original optimization objective (e.g. $L(s\cdot\lfloor w / s \rceil), w\in \mathbb{R}^{d}, s\in\mathbb{R}$) with a \textbf{biased} objective (e.g. $E_{\xi_i \sim U(-0.5, 0.5), \forall i}[L(s\cdot(w/s+\xi))], w \in \mathbb{R}^d, s\in\mathbb{R}, \xi \in \mathbb{R}^d$). Sections~\ref{sec:eval_toy} and~\ref{sec:pqn} empirically show that PQN cannot consistently outperform baseline QAT methods like the STE, and we hypothesize this is due to the biasedness of the optimization objective. Sections~\ref{sec:eval_toy} and \ref{sec:pqn_exp} empirically and theoretically demonstrate this biasedness for a toy problem. Sections~\ref{sec:eval_toy} and~\ref{sec:pqn} show that CEWT consistently outperforms both PQN and the STE. 

Existing methods demonstrates that reducing the rate of weight oscillation leads to significant accuracy improvement, but tuning hyperparameter is non-trivial and a tuned hyperparameter for a model size/architecture might not transfer to another. On the other hand, existing hyperparameter-free methods like PQN cannot consistently outperform the STE. We therefore ask the research question: \textit{can we regulate the rate of rounding boundary weight oscillation and consistently achieve higher-than-STE accuracy without introducing any hyperparameters}? 
As a preview, CEWT regulates the number of parameters near rounding boundaries, without introducing any hyperparameters, by enforcing the empirical weight distribution to match a zero-mean Gaussian, an assumption embedded in the design of many quantizers~\citep{yang2026boosting,panferov2025queststabletrainingllms,dettmers2023qloraefficientfinetuningquantized} but is not enforced in practice.

\subsection{Hypersphere Optimization}\label{sec:ho}
Hypersphere Optimization (HO)~\citep{xie2026controlled,hyperballOptimization2026,ren2026rethinkinglanguagemodelscaling} is a family of optimizers that \textit{removes the hyperparameter weight decay} by constraining weights to lie on the radius-$R$ Frobenius norm hypersphere~\citep{hyperballOptimization2026} or the radius-$R$ Spectral norm hypersphere~\citep{xie2026controlled}. HO targets the following problem:
\begin{equation}\label{eqn:hyperball_problem}
    \underset{W \in \mathbb{R}^{m \times n}}{\text{min. }} L(W) \text{ s.t. } ||W|| = R
\end{equation}
where $L$ denotes the loss as a function of the parameters, $W \in \mathbb{R}^{m \times n}$ is the weight matrix of a linear layer\footnote{Without loss of generality, we assume there is only one layer in the model and it is a linear one.} with $m$ output features and $n$ input features, $||\cdot||$ is either the Frobenius norm operator $||\cdot||_F$ or the spectral norm operator $||\cdot||_2$, and $R\in \mathbb{R}$ is typically chosen to be $||W^{(0)}||$, $W$'s norm at random initialization time~\citep{hyperballOptimization2026,ren2026rethinkinglanguagemodelscaling}. The algorithm of HO is as follows:
\begin{subequations}\label{eqn:hyperball}
\begin{align}
    \bar{W}^{(t)} &= W^{(t-1)} - \eta^{(t)} R U^{(t)} \label{eqn:hyperball_a}\\
    \hat{W}^{(t)} &= \bar{W}^{(t)} \label{eqn:hyperball_b}\\
    W^{(t)} &= \frac{R}{||\hat{W}^{(t)}||}\hat{W}^{(t)} \label{eqn:hyperball_c} 
\end{align}
\end{subequations}
where $\cdot^{(t)}$ is the value of a variable at iteration $t$, $U \in \mathbb{R}^{m \times n}$ is a unit-norm (i.e. $||U||=1$) update whose definition is optimizer-dependent (e.g. see Algorithms~\ref{alg:adamh}, \ref{alg:muonh}, \ref{alg:lionh}, and \ref{alg:sso}), and $\bar{W}, \hat{W} \in \mathbb{R}^{m \times n}$ are variables introduced for notational convenience. 
In Equation~\ref{eqn:hyperball_a}, HO scales the update $U$ to have norm $R$ and performs the update with a learning rate $\eta$, ensuring the ratio $||\Delta W||/||W|| $ is exactly $\eta$, a key property that enables hyperparameter (learning rate) transfer~\citep{hyperballOptimization2026,ren2026rethinkinglanguagemodelscaling,yang2024spectralconditionfeaturelearning}. Equation~\ref{eqn:hyperball_b} is a redundant equation that we intentionally introduced, and we will return to discuss Equation~\ref{eqn:hyperball_b} in Section~\ref{sec:cewt_alg}. In Equation~\ref{eqn:hyperball_c}, HO projects the updated weight $\hat{W}$ back to the radius-$R$ hypersphere. Hypersphere optimizers (e.g. AdamH/MuonH/SSO) empirically outperform~\citep{xie2026controlled,hyperballOptimization2026,ren2026rethinkinglanguagemodelscaling} non-hypersphere optimizers (e.g. Adam/Muon). 

By converting a \textbf{soft constraint} (e.g. penalize the Frobenius norm of weights as much as possible) into a \textbf{hard constraint} (e.g. the Frobenius norm of weights must be exactly $R$), HO successfully removes the need to tune hyperparameters, as there exist plenty of theoretically-grounded methods~\citep{yang2024spectralconditionfeaturelearning,pmlr-v139-yang21c,NEURIPS2021_8df7c2e3} to choose\footnote{Section~\ref{sec:sweep_r} discusses that $R$ is \textbf{not} introduced by HO, but is inherited from weight initialization methods and \textbf{reused} by HO.} $R$, whereas choosing the weight decay hyperparameter generally relies on empirical observations and expensive grid searches.
Therefore, if it is easier to describe \textbf{what a ``good'' solution looks like}, compared to \textbf{what regularization term should be added to the loss such that the model is ``good'' at the end of training}, hard constraints are an effective tool to remove the less expressive and harder-to-tune soft constraint hyperparameters. Returning to the problem of hyperparameter-free oscillation suppression in QAPT, we then ask the question: \textit{what does the histogram of the weights of a ``good'' , or oscillation-free, model look like?}

\section{Proposed Method}\label{sec:proposed}
In this section, we present CEWT. 
CEWT is designed based on the observation that the weight distribution of round-boundary-oscillation-free models, specifically those trained \textit{without any quantization} (rounding boundary oscillation is a QAT-specific phenomenon), tend to look like a zero-mean Gaussian~\citep{dettmers2023qloraefficientfinetuningquantized}. Therefore, many quantizer design decisions, such as choosing clipping functions~\citep{yang2026boosting}, scaling factors~\citep{panferov2025queststabletrainingllms}, and bin centers~\citep{dettmers2023qloraefficientfinetuningquantized}, are made based on the zero-mean Gaussian assumption.
However, during QAT, the histogram of pre-clip\footnote{The definition of pre-clip weights vary across quantizers. See Algorithms~\ref{alg:bbq}, \ref{alg:quest}, and \ref{alg:lsq} for a few examples.} weights do \textit{not} look like zero-mean Gaussians due to weight density spikes that appear at rounding boundaries (see Figure~\ref{fig:intro} for an example) . To address this gap between assumption and reality, we propose to encode the zero-mean Gaussian prior as a hard constraint (Equation~\ref{eqn:cewt_problem_c}), thereby constraining the entire optimization trajectory in density-spike-free regions of the weight space. CEWT targets the following optimization problem:
\begin{subequations}\label{eq:cewt_problem}
\begin{align}
\underset{W \in \mathbb{R}^{m \times n}}{\text{min. }} & L(Q) \\
\text{ s.t. } &Q = \textit{weight\_quantizer}(W) \label{eqn:cewt_problem_a}\\
& W = \frac{R}{||\hat{W}||} \hat{W} \label{eqn:cewt_problem_b}\\
&\text{sort}(\text{vec}(\hat{W}))_i = \Phi^{-1}\left(\frac{i+0.5}{mn}\right), \forall i \in \{0, 1, \cdots, mn-1\} \label{eqn:cewt_problem_c}
\end{align}
\end{subequations}
where $Q \in D_Q \subsetneq \mathbb{R}^{m \times n}$ denotes the quantized parameters of a linear layer\footnote{Without loss of generality, we assume there is only one layer in the model and it is a linear one.}; $L: D_Q \rightarrow \mathbb{R}$ denotes the loss as a function of quantized parameters and typically contains, as part of $L$'s definition, the model architecture, dataset, and \textbf{activation quantizers}; \textit{weight\_quantizer}$:\mathbb{R}^{m \times n}\rightarrow D_Q$ describes the behaviour of a generic quantizer like BBQ, QuEST, or LSQ; $\Phi^{-1}:(0, 1)\rightarrow\mathbb{R}$ is the inverse function of the standard Gaussian CDF; and sort$: \mathbb{R}^{mn}\rightarrow \mathbb{R}^{mn}$ treats its input vector as an array of numbers and returns the sorted list. Equation~\ref{eqn:cewt_problem_a} is a standard constraint for QAT. Equation~\ref{eqn:cewt_problem_b} states that CEWT is an extension to HO (Section~\ref{sec:ho}), as a $W$ that satisfy Equation~\ref{eqn:cewt_problem_b} must lie on the radius-$R$ (Frobenius/spectral norm) hypersphere. Equation~\ref{eqn:cewt_problem_c} is the formal description of the zero-mean Gaussian assumption: for a sufficiently large $mn$, a random permutation of $\text{vec}(\hat{W})$ must look indistinguishable from a list of $mn$ samples i.i.d. sampled from $N(0, 1)$. In essence, Equation~\ref{eqn:cewt_problem_c} constrains the \textbf{center} (zero) and \textbf{shape} (Gaussian) of the weight distribution, while Equation~\ref{eqn:cewt_problem_b} constrains the \textbf{width} of the weight distribution.

We discuss the algorithm of CEWT in Section~\ref{sec:cewt_alg}. In Sections~\ref{sec:cewt_ht} and~\ref{sec:cewt_tgcs}, we discuss how the function \textit{weight\_quantizer} in Equation~\ref{eqn:cewt_problem_a} can be customized to maximally utilize the oscillation suppression effect of CEWT, 

\subsection{Algorithm}\label{sec:cewt_alg}
\begin{figure}[t]
    \centering
    \begin{subfigure}[t]{0.245\textwidth}\hfill
        \includegraphics[width=\linewidth]{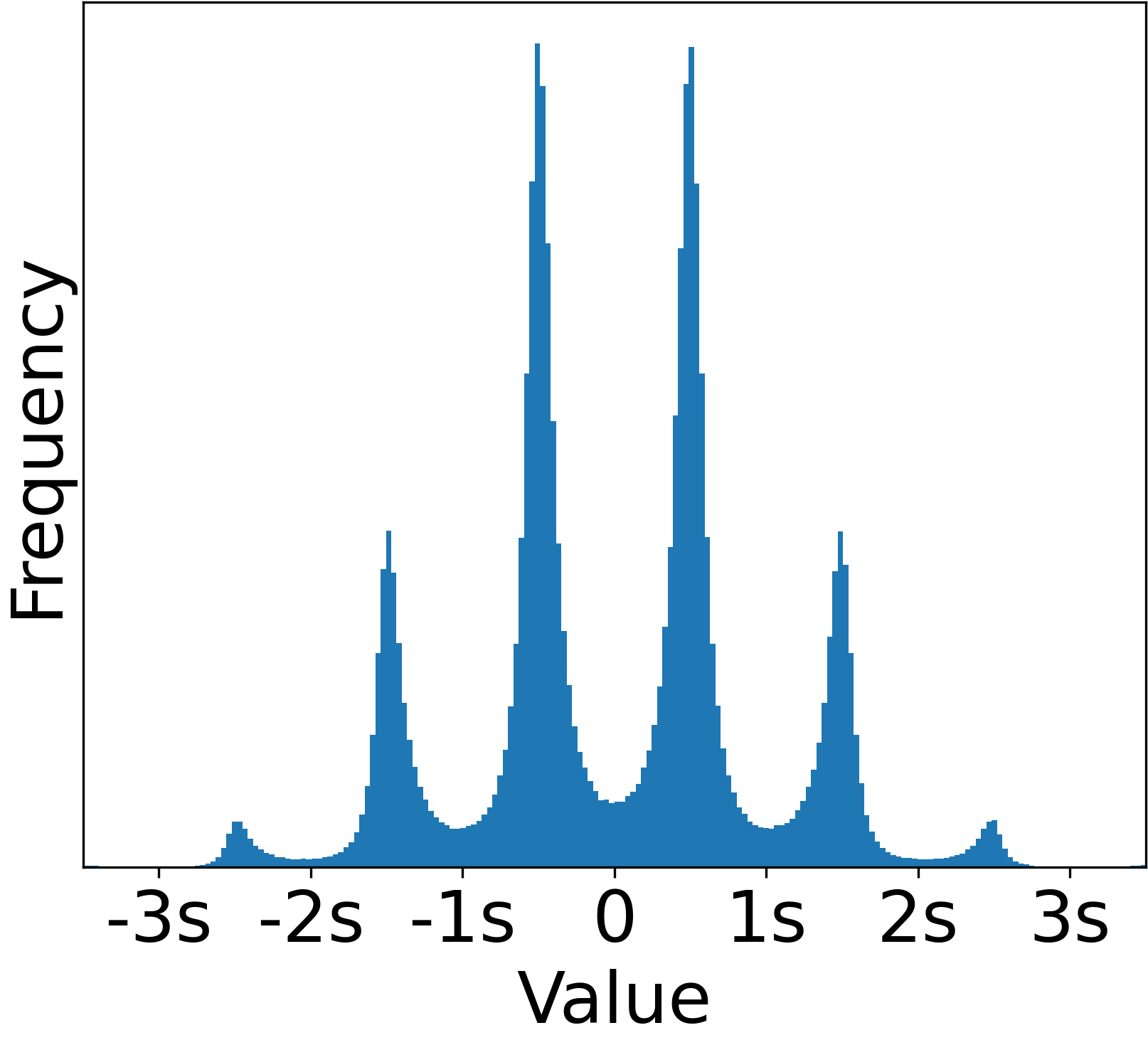}
        \caption{Example $\bar{W}$}
        \label{fig:cewt1}
    \end{subfigure}
    \begin{subfigure}[t]{0.245\textwidth}\hfill
        \includegraphics[width=\linewidth]{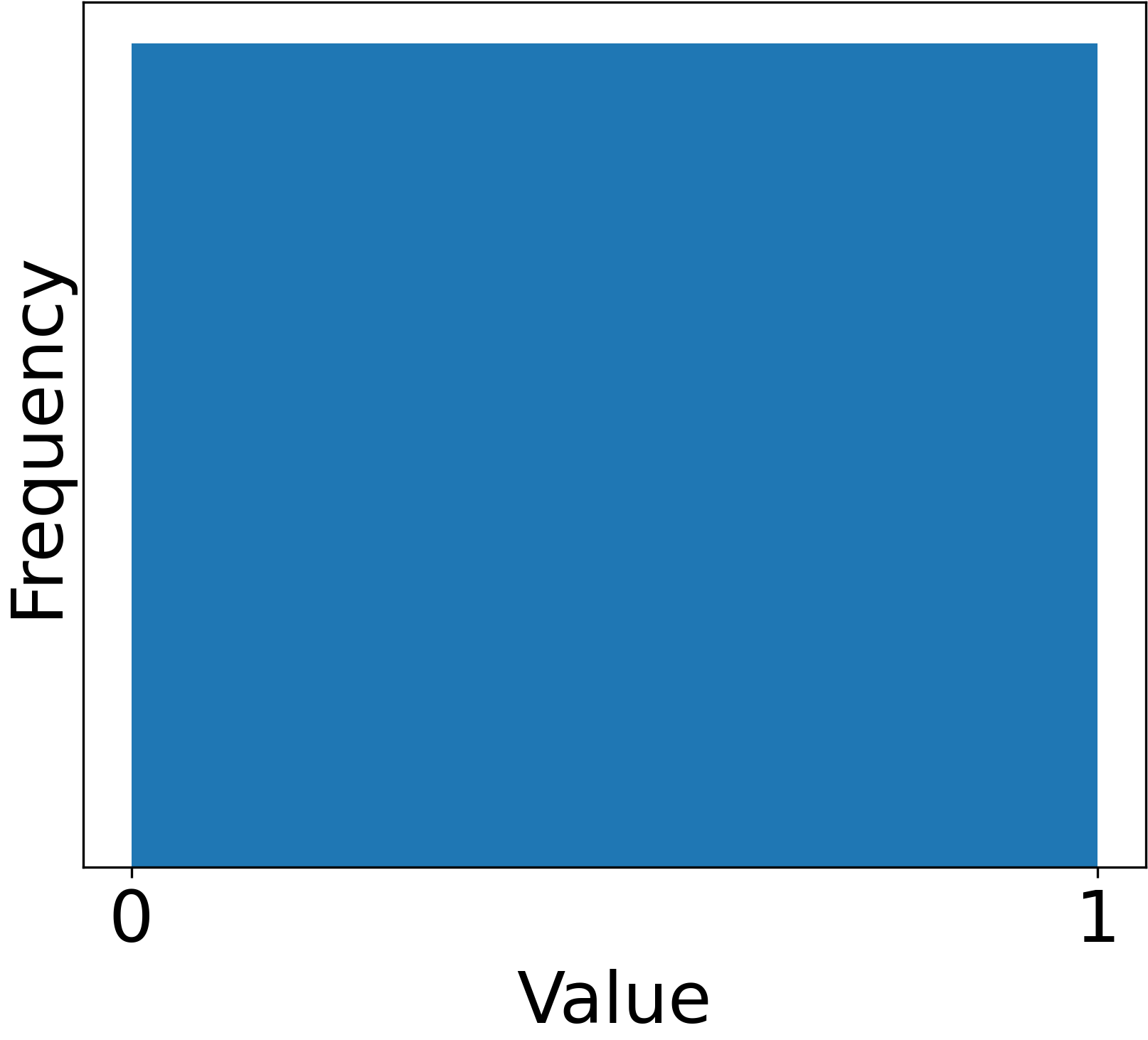}
        \caption{$\frac{1}{mn}\cdot\left(\text{ranks}(\bar{W}) + \frac{1}{2}\right)$}
        \label{fig:cewt2}
    \end{subfigure}
    \begin{subfigure}[t]{0.245\textwidth}\hfill
        \includegraphics[width=\linewidth]{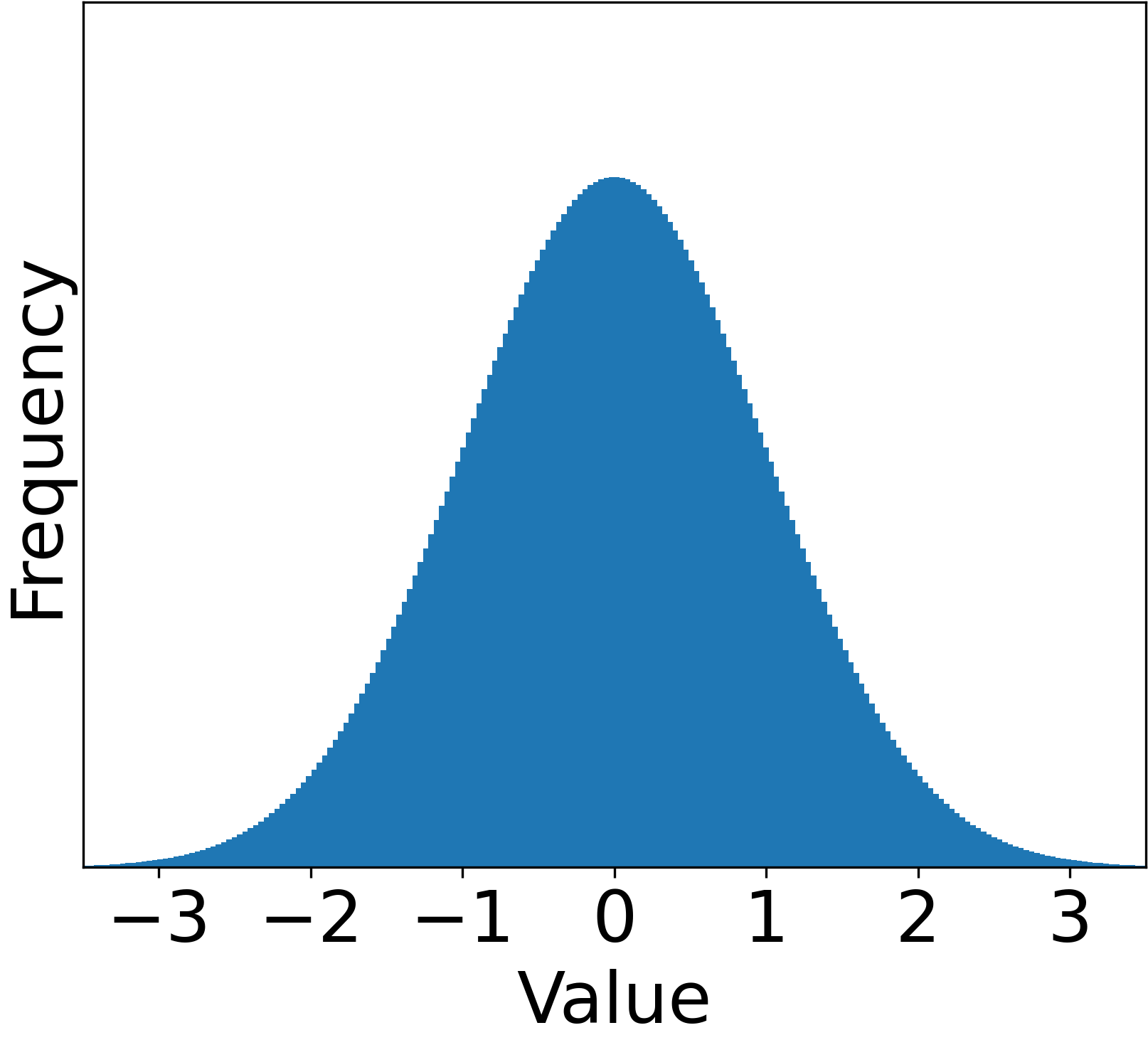}
        \caption{$\hat{W}$}
        \label{fig:cewt3}
    \end{subfigure}
    \begin{subfigure}[t]{0.245\textwidth}\hfill
        \includegraphics[width=\linewidth]{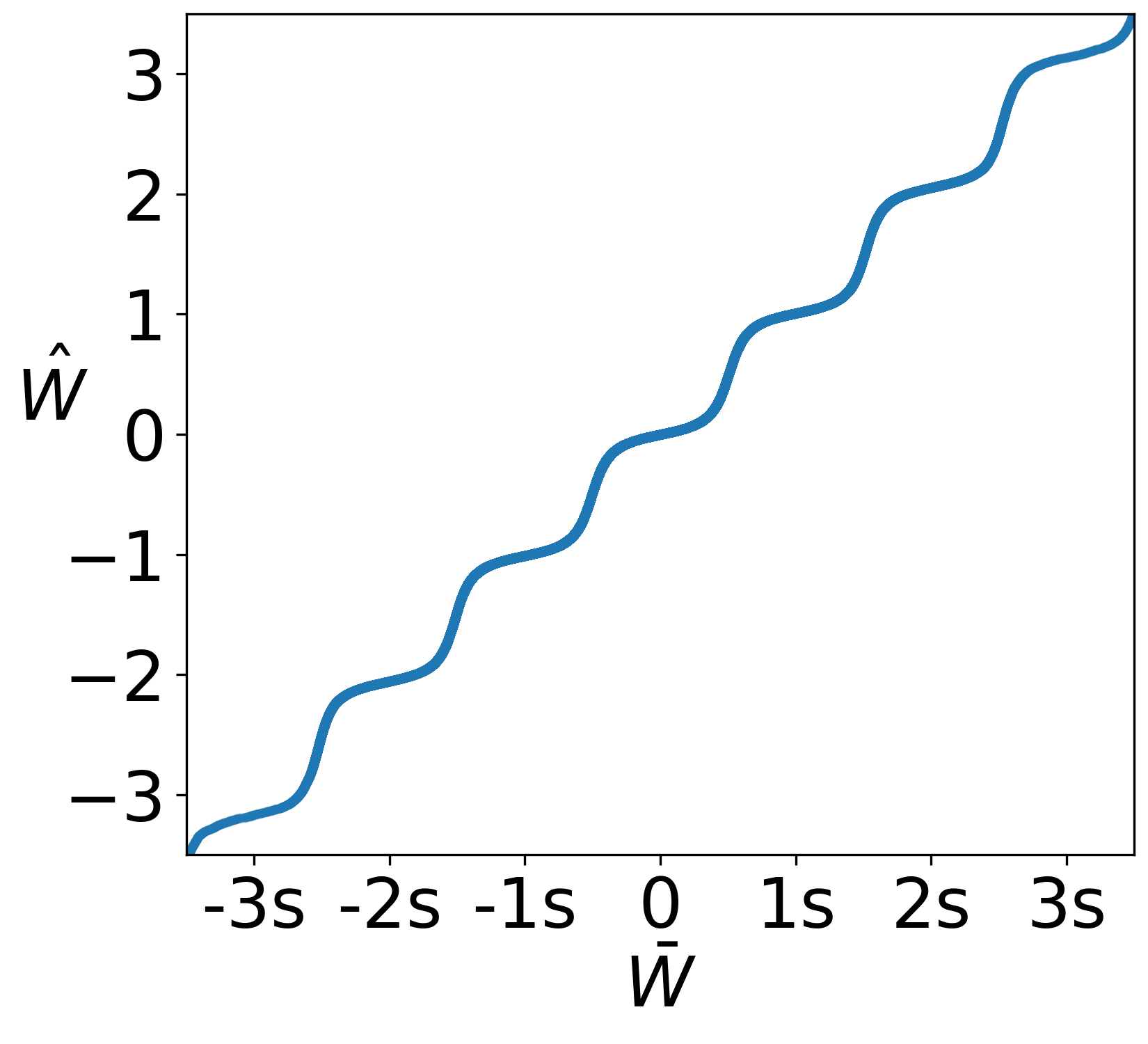}
        \caption{CEWT's mapping}
        \label{fig:cewt4}
    \end{subfigure}
    \caption{Visualization of CEWT (Equation~\ref{eqn:cewt}). Notations like vec$(\cdot)$ and $\cdot^{(t)}$ are omitted. The example $\bar{W}$ we use here is the weights of the STE experiment in Table~\ref{tab:toy} at iteration 60. The $s$ in sub-figures~\ref{fig:cewt1} and \ref{fig:cewt4} is defined in Section~\ref{sec:toy_details}.}
    \label{fig:cewt}
\end{figure}

The algorithm of CEWT is Equation~\ref{eqn:cewt}, which is meant to \textbf{replace Equation~\ref{eqn:hyperball_b}} as CEWT is an extension to hypersphere optimizers. Figure~\ref{fig:cewt} visualizes Equation~\ref{eqn:cewt}.
\begin{equation}\label{eqn:cewt}
    \hat{W}^{(t)} = \text{mat}\left(\Phi^{-1}\left(\frac{1}{mn}\cdot\left(\text{ranks}(\text{vec}(\bar{W}^{(t)})) + \frac{1}{2}\right)\right)\right)
\end{equation}
where $\Phi^{-1}$ is the element-wise inverse standard Gaussian CDF operation, vec$:\mathbb{R}^{m \times n} \rightarrow \mathbb{R}^{mn}$ flattens a matrix into a vector, mat$: \mathbb{R}^{mn} \rightarrow \mathbb{R}^{m \times n}$ is the reverse operation (un-flattening), and ranks$:\mathbb{R}^{mn} \rightarrow \{0, 1, \cdots mn-1\}^{mn}$ is defined as:
\begin{equation}\label{eqn:rank}
\begin{split}
    \text{ranks}(w)_i=\sum_{k=0}^{mn-1} \mathbf{1}_{(w_k < w_i) \lor (w_k = w_i \land k < i)}, \forall i \in \{0, 1, \cdots, mn-1 \}
\end{split}
\end{equation}
where $\mathbf{1}_{\text{cond}}$ returns a one if cond is true and a zero if cond is false. The function ranks counts the number of elements ($w_k$) in the input list ($w$) that is smaller than the current element ($w_i$).
If multiple elements in $w$ are equivalent ($w_k = w_i$), we use which element comes first in the list ($k < i$) to break ties, ensuring \textbf{each element is assigned a unique rank}. After ranking, the ranks are normalized to the range $(0, 1)$. Note the histogram of normalized ranks (visualized in Figure~\ref{fig:cewt2}) is guaranteed to be uniform as each rank shows up only once. Next, the normalized ranks are projected to Gaussian quantiles using the inverse standard Gaussian CDF. Since the normalized ranks are uniform, the histogram of projected quantiles (visualized Figure~\ref{fig:cewt3}) is guaranteed to be standard Gaussian (known as inverse transform sampling). Algorithm~\ref{alg:cewt} shows a PyTorch-like implementation of Equation~\ref{eqn:cewt}. Note that ranks can be efficiently implemented with argsort and scatter.

\begin{algorithm}[t]
\caption{PyTorch-like Implementation of CEWT (Equation~\ref{eqn:cewt})}
\label{alg:cewt}
\begin{algorithmic}[1]
    \Function{CEWT}{wbar} 
        \State indices = argsort(wbar.flatten()) 
        \State ranks = zeros\_like(indices)
        \State ranks[indices] = arange(indices.numel()) {\color{gray} \# scatter}
        \State ranks = (ranks.to(float64) + 0.5) / ranks.numel() {\color{gray} \# use float64 for large $mn$}
        \State what = erfinv(2 * ranks - 1) * (2 ** 0.5) {\color{gray} \# $\Phi^{-1}(x)=\sqrt{2}\cdot\text{erf}^{-1}(2x-1)$}
        \State return what.to(float32).view\_as(wbar)
    \EndFunction
\end{algorithmic}
\end{algorithm}

Equation~\ref{eqn:cewt} has a few properties. First, relative order is preserved ($\bar{W}_{ij} \le \bar{W}_{kl} \implies \hat{W}_{ij} \le \hat{W}_{kl}$) because each operation (ranks, multiplication by positive scalar, addition by positive scalar, and inverse standard Gaussian CDF) preserves relative order. For example, Figure~\ref{fig:cewt4} shows the mapping from an example $\bar{W}$ to its corresponding $\hat{W}$ is a non-decreasing function. Second, the histogram of $\hat{W}$ is always the standard Gaussian (as noted above) and therefore $\hat{W}$ satisfies Constraint~\ref{eqn:cewt_problem_c}. This zero-mean Gaussian property can be exploited by quantizers to achieve desirable properties such as quantization error minimization~\citep{panferov2025queststabletrainingllms} and entropy maximization~\citep{yang2026boosting}. Third, if the histogram of $\bar{W}$ contains rounding boundary density spikes (e.g. Figure~\ref{fig:cewt1}), CEWT's mapping effectively pulls weights too close to rounding boundary away from rounding boundary (e.g. Figure~\ref{fig:cewt4}). As such, CEWT can be interpreted as a hyperparameter-free hard-constraint variant of Dampening Loss~\citep{pmlr-v162-nagel22a} or Periodic Interpolation~\citep{li2026winqacceleratingquantizationawaretraining}. 

\subsection{Removing Hadamard Transforms}\label{sec:cewt_ht}
Many implementations of \textit{weight\_quantizer} include orthogonal transforms such as the Hadamard Transform~\citep{yang2026boosting,panferov2025queststabletrainingllms,lin2024duquantdistributingoutliersdual}, and we discuss two options to integrate such implementations with CEWT and why we choose the second option. The first option is to keep the original implementation of \textit{weight\_quantizer} and enforce the zero-mean Gaussian constraint in the Hadamard domain by converting $\bar{W}^{(t)}$ to the Hadamard domain, applying vanilla CEWT (Equation~\ref{eqn:cewt}), and subsequently projecting the results back to the original domain, resulting in Equation~\ref{eqn:cewt_ht}:
\begin{equation}\label{eqn:cewt_ht}
    \hat{W}^{(t)} = \textbf{HT}\left(\text{mat}\left(\Phi^{-1}\left(\frac{1}{mn}\cdot\left(\text{ranks}(\text{vec}(\textbf{HT}(\bar{W}^{(t)})) + \frac{1}{2}\right)\right)\right)\right)
\end{equation}
The second option is to \textbf{remove} Hadamard Transform (HT) from the implementation of \textit{weight\_quantizer} and use \textbf{vanilla} CEWT (Equation~\ref{eqn:cewt}). In terms of computational cost, the second option saves four Hadamard Transforms per layer per training iteration: one less HT in the forward pass of \textit{weight\_quantizer}, one less HT in the backward pass, and two less HTs in the optimizer (Equation~\ref{eqn:cewt} compared to \ref{eqn:cewt_ht}). In terms of empirical performance, Section~\ref{sec:eval_deep} shows that the second option consistently achieves lower loss than the first one. Therefore, when we use CEWT, we \textbf{remove the Hadamard Transform from \textit{weight\_quantizer}}.

\subsection{Tensor-Wise Quantization Grid with Channel-wise Scaling (TGCS)}\label{sec:cewt_tgcs}
\begin{table}[]
\centering
\resizebox{\textwidth}{!}{%
\renewcommand{\arraystretch}{1.3} 
\begin{tabular}{|c|c|c|}
\hline
Per-Tensor & Per-Channel & TGCS (Ours) \\ \hline
{\color{orange!50!black} $s_q = \text{quant\_scale}(\text{vec}(W))$}           &  {\color{blue!50!black} $s_{qi} = \text{quant\_scale}(W_{i})$}     &  {\color{orange!50!black}$s_q = \text{quant\_scale}(\text{vec}(W))$} \\ 
{\color{orange!50!black} $s_d = \text{dequant\_scale}(\text{vec}(W))$}           & {\color{blue!50!black}$s_{di} = \text{dequant\_scale}(W_{i})$}     & {\color{blue!50!black}$s_{di} = \text{dequant\_scale}(W_{i})$} \\ 
$Q = {\color{orange!50!black}s_d}\cdot \text{clipround}(1 / {\color{orange!50!black}s_q} \cdot W)$           &  $\forall i, Q_{i} = {\color{blue!50!black}s_{di}}\cdot \text{clipround}(1 / {\color{blue!50!black}s_{qi}} \cdot W_{i})$        & $\forall i, Q_{i} = {\color{blue!50!black}s_{di}}\cdot \text{clipround}(1 / {\color{orange!50!black}s_{q}} \cdot W_{i})$   \\ \hline
\end{tabular}%
}
\caption{Comparison between per-tensor, per-channel, and TGCS quantization. Example definitions of quant\_scale, dequant\_scale, and clipround are found in Algorithms~\ref{alg:bbq}, \ref{alg:quest}, and \ref{alg:lsq}.}
\label{tab:tgcs}
\end{table}
Many implementations of \textit{weight\_quantizer} uses \textbf{per-channel} quantization as opposed to \textbf{per-tensor} quantization (see Table~\ref{tab:tgcs} for the difference), and we discuss how to integrate CEWT with such per-channel weight quantizers.
The underlying assumption of per-channel quantizers like BBQ (without HT) and QuEST (without HT) is that every \textbf{individual channel} (e.g. $W_{0}\in \mathbb{R}^n$, $W_{1}\in \mathbb{R}^n$, etc.) is a permutation of zero-mean Gaussian noise, whereas Equation~\ref{eqn:cewt} only enforce the zero-mean Gaussian constraint at the \textbf{tensor} level.
To address this mismatch, we propose TGCS (see Table~\ref{tab:tgcs}) which decouples the granularity between \textbf{quantization} and \textbf{dequantization}. Specifically, TGCS uses a \textbf{per-tensor quantization} scaling factor to construct a tensor-wise quantization grid whose underlying assumption is enforced by Equation~\ref{eqn:cewt}, and \textbf{per-channel de-quantization} scaling factors to achieve the performance benefit of per-channel quantization. Section~\ref{sec:eval_deep} visually (Figure~\ref{fig:ablation}) and quantitatively (Table~\ref{tab:ablation}) show that TGCS improves the oscillation suppression effect of CEWT relative to per-channel quantization. To summarize, when we use CEWT, we \textbf{modify \textit{weight\_quantizer} to use TGCS if the original implementation is per-channel}.

\section{Evaluation}
In this section, we systematically evaluate the effectiveness of CEWT. In Section~\ref{sec:eval_toy}, we compare CEWT against prior oscillation suppression methods, evaluated on a simple but important toy problem. In Section~\ref{sec:eval_deep}, we then evaluate the effectiveness of CEWT as a hyperparameter-free extension to existing hypersphere optimizers, by pre-training deeper networks under quantization. In Section~\ref{sec:eval_deep}, we also perform ablation studies to verify the insights in Sections~\ref{sec:cewt_ht} and \ref{sec:cewt_tgcs}.
\subsection{Toy Problem}\label{sec:eval_toy}
In this section, we evaluate the effectiveness of CEWT on a representative toy problem and confirm CEWT can, without introducing any hyperparameters, reach performance comparable to prior oscillation suppression methods. Our toy problem is as follows:
\begin{equation}\label{eqn:toy}
    \begin{split}
        \underset{W}{\text{min}} \text{ } \frac{1}{2} ||XQ - XW^*||_F^2 
        \text{ s.t. } Q = s\cdot \lfloor 1 / s \cdot W \rceil \text{ and } ||W||_F = ||W^*||_F
    \end{split}
\end{equation}
where $X \in \mathbb{R}^{n\times d}$ is a dataset with $n$ data points each with $d$ features, $W \in \mathbb{R}^{d\times d}$ is the raw weights, $W^*\in \mathbb{R}^{d\times d}$ is one (not necessarily unique) optimal solution when no quantization (rounding) is involved, $XW^* \in \mathbb{R}^{n\times d}$ is the regression targets (expected output), $\lfloor \cdot \rceil$ is the element-wise rounding operation, and $s\in \mathbb{R}$ is the quantization step size. 
Section~\ref{sec:toy_expl} discusses that Problem~\ref{eqn:toy} is a representative problem because it simulates multiple phenomena in the QAT of deep networks.

We compare CEWT against prior oscillation suppression methods: Pseudo-Quantization Noise (PQN)~\citep{Shin_2023_CVPR}, Dampening Loss (DL)~\citep{pmlr-v162-nagel22a}, Iterative Freezing (IF)~\citep{pmlr-v162-nagel22a}, Confidence-Guided Annealing (CGA)~\citep{10.5555/3618408.3619311}, and Periodic Interpolation (PI)~\citep{li2026winqacceleratingquantizationawaretraining}. Section~\ref{sec:toy_details} shows details on how these algorithms are applied to Problem~\ref{eqn:toy}. If an oscillation suppression method comes with a hyperparameter, we choose a good hyperparameter and a bad hyperparameter, and report the mean-squared error between $XQ$ and $XW^*$, and between $XW$ and $XW^*$. To quantify weight oscillation, we report EMA Oscillation Frequency~\citep{pmlr-v162-nagel22a} and Rounding Boundary Mass (RBM) which is inspired by \citet{10.5555/3618408.3619311}. Section~\ref{sec:quantify_wo} discusses the details of these two metrics and their tradeoffs. Table~\ref{tab:toy} suggests while existing non-PQN oscillation-suppression methods can consistently suppress rounding boundary spikes, the MSE they can achieve is highly dependent on the chosen hyperparameter. In contrast, CEWT can, without the need for tuning, achieve performance comparable to non-PQN techniques with well-tuned hyperparameters. While PQN is hyperparameter-free and effective in oscillation suppression, it underperforms compared to the STE and CEWT due to the biasedness\footnote{Section~\ref{sec:pqn_exp} discusses why MSE$(XW, XW^*)$ is PQN's true, yet biased, objective.} in the optimization objective---PQN excels in minimizing MSE$(XW, XW^*)$, but the STE and CEWT can find $W$ that performs better in terms of the true objective MSE$(XQ, XW^*)$. Since oscillation happens regardless~\citep{pmlr-v162-nagel22a} of the rounding Jacobian estimation method, we also evaluate the effectiveness of CEWT when applied to HTGE~\citep{10227741}, EWGS~\citep{Lee_2021_CVPR}, and JacQuant-Probe~\citep{yi2026jacquantstefreequantizationawaretraining} and show our results in Table~\ref{tab:toy_grad_est}. Results suggest CEWT can improve performance and suppress oscillation \textbf{regardless of the rounding Jacobian estimator}.
\begin{table}[]
\centering
\resizebox{\textwidth}{!}{%
\begin{tabular}{|l|l|c|c|c|c|c|c|c|}
\hline
                            &         & STE                 & PQN  & CEWT                            & DL & IF & CGA        & PI \\ \hline
\multirow{2}{*}{MSE$(XQ, XW^*)$ (unit: 1e-3)} & Good HP &\multirow{2}{*}{81±0} & \multirow{2}{*}{87±0} & \multirow{2}{*}{\textbf{45±0}} & \textbf{30±0} & 59±0              & 53±0      & 54±0                  \\ \cline{2-2} \cline{6-9} 
                            & Bad HP  &     &                   &                                 & 375±1          & 2080±12              & \textbf{332±1}      & 1847±10                  \\ \hline
\multirow{2}{*}{MSE$(XW, XW^*)$ (unit: 1e-3)} & Good HP &\multirow{2}{*}{67±0} & \multirow{2}{*}{\textbf{4±0}}  & \multirow{2}{*}{36±0} & \textbf{34±0} & 60±0              & 55±0      & 48±0                  \\ \cline{2-2} \cline{6-9} 
                            & Bad HP  &        &                &                                 & \textbf{295±1}          & 2080±12              & 386±1       & 1807±9                  \\ \hline
\multirow{2}{*}{EMA Oscillation Freq (unit: 1e-4)}        & Good HP & \multirow{2}{*}{701±2} & \multirow{2}{*}{\textbf{9±0}}  & \multirow{2}{*}{17±0} & 5±0      & 20±0              & 264±5      & \textbf{0±0}             \\ \cline{2-2} \cline{6-9} 
                            & Bad HP  &          &              &                                 & \textbf{0±0}     & \textbf{0±0}         & 1±0 & \textbf{0±0}             \\ \hline
\multirow{2}{*}{RBM$(\text{vec}(1/s\cdot W))$ (unit: 1e-3)}        & Good HP & \multirow{2}{*}{573±1} & \multirow{2}{*}{\textbf{10±0}}  & \multirow{2}{*}{\textbf{10±0}} & 6±0      & 13±0              & 133±3      & \textbf{0±0}             \\ \cline{2-2} \cline{6-9} 
                            & Bad HP  &          &              &                                 & \textbf{0±0}     & \textbf{0±0}         & \textbf{0±0} & \textbf{0±0}             \\ \hline
\end{tabular}%
}
\caption{Comparison between weight oscillation suppression methods. We report mean plus minus 2 standard deviations over 5 random seeds. Section~\ref{sec:quantify_wo} defines EMA Oscillation Frequency and RBM.}
\label{tab:toy}
\end{table}

\subsection{Deep Learning Experiments and Ablation Studies}\label{sec:eval_deep}
In this section, we evaluate the effectiveness of CEWT (Equation~\ref{eqn:cewt}, Section~\ref{sec:cewt_ht}, and Section~\ref{sec:cewt_tgcs}) as a hyperparameter-free extension to existing hypersphere optimizers. We perform Quantization-Aware \text{Pre-training} with various combinations of two model architectures (LLaMA~\citep{touvron2023llamaopenefficientfoundation} and GPT~\citep{Radford2019LanguageMA}), 
various model scales (95M, 125M, 200M, 300M, 600M), four optimizers (AdamH~\citep{kingma2017adammethodstochasticoptimization,hyperballOptimization2026}, MuonH~\citep{muon,hyperballOptimization2026}, LionH~\citep{chen2023symbolicdiscoveryoptimizationalgorithms,hyperballOptimization2026}, and SSO~\citep{xie2026controlled}), two precisions (1-bit and 2-bit), and three quantizers (LSQ, QuEST, and BBQ). Sections~\ref{sec:optimizer} and \ref{sec:quantizer} show detailed algorithms for optimizers and quantizers. Section~\ref{sec:full} presents our learning rates and full-precision model perplexity. Our training recipe large follows~\citet{panferov2025queststabletrainingllms}, with 100 tokens per non-embedding parameter, significantly exceeding the Chinchilla scaling~\citep{hoffmann2022trainingcomputeoptimallargelanguage}. In total, our experiments (in Table~\ref{tab:main}) consumed \textbf{30 RTX-5090-days and 34 H100-days}. 
For every experiment, we quantize both activations and weights of linear layers.  We present pre-training perplexity in Table~\ref{tab:main}, zero-shot evaluation in Section~\ref{sec:zero_shot}, and training time in Section~\ref{sec:train_time}. Results suggest that with a geomean increase of 4\% in training time, \textbf{CEWT can consistently improve the performance of hypersphere optimizers in QAPT without introducing any hyperparameters}. Unlike the RBM of HO, the RBM of CEWT is \textbf{invariant} across optimizers and model sizes, and the RBM of CEWT depends \textbf{entirely} on the quantizer and precision\footnote{For BBQ, since pre-rounding weights are uniformly distributed, the RBM of CEWT is 0.010 regardless of precision. For QuEST and LSQ, the RBM of CEWT depends on the precision. }. This invariance is a \textit{key} feature of CEWT---ensuring the number of parameters near rounding boundary is a constant. Figure~\ref{fig:ema_osc_freq_bbq} reports \textbf{both} EMA Oscillation Frequency and RBM of a 2-bit LLaMA-95M during training, suggesting that the perplexity reduction of CEWT comes from its oscillation suppression effect.  Section~\ref{sec:vision} shows CEWT improves the accuracy of vision models. Section~\ref{sec:pi_dl_cewt} shows CEWT can match/outperform the performance of Periodic Interpolation and Dampening Loss with the best hyperparameter tuned using grid search, and Section~\ref{sec:pqn} shows CEWT consistently outperforms the hyperparameter-free method PQN.

We additionally perform ablation studies to test the effects of removing the HT and TGCS. We compare HO, CEWT without TGCS, CEWT, and CEWT-HT (first option in Section~\ref{sec:cewt_ht}). Table~\ref{tab:ablation} and Figure~\ref{fig:ablation} suggest that TGCS can improve the density spike suppression effect of CEWT. Table~\ref{tab:ablation} suggests that the combination of removing the HT from \textit{weight\_quantizer} and using vanilla CEWT outperforms the combination of keeping the HT  and using CEWT-HT. Section~\ref{sec:sweep_r} shows that the perplexity reduction of CEWT compared to HO is consistent regardless of $R$.  
\begin{table}[!htbp]
\centering
\resizebox{\textwidth}{!}{%
\begin{tabular}{|l|c|cccc|cccc|cccc|cccc|}
\hline
\multicolumn{1}{|c|}{\multirow{2}{*}{Model}} & \multirow{2}{*}{Q-B} & \multicolumn{2}{c|}{MuonH} & \multicolumn{2}{c|}{Muon-CEWT} & \multicolumn{2}{c|}{AdamH} & \multicolumn{2}{c|}{Adam-CEWT} & \multicolumn{2}{c|}{LionH} & \multicolumn{2}{c|}{Lion-CEWT}  & \multicolumn{2}{c|}{SSO} & \multicolumn{2}{c|}{SSO-CEWT}  \\
& &\multicolumn{1}{c|}{PPL} & \multicolumn{1}{c|}{RBM} &\multicolumn{1}{c|}{PPL} & \multicolumn{1}{c|}{RBM} &\multicolumn{1}{c|}{PPL} & \multicolumn{1}{c|}{RBM} &\multicolumn{1}{c|}{PPL} & \multicolumn{1}{c|}{RBM} &\multicolumn{1}{c|}{PPL} & \multicolumn{1}{c|}{RBM} & \multicolumn{1}{c|}{PPL} & \multicolumn{1}{c|}{RBM} & \multicolumn{1}{c|}{PPL} & \multicolumn{1}{c|}{RBM}  & \multicolumn{1}{c|}{PPL} & \multicolumn{1}{c|}{RBM} \\ \hline
L-95  & L-1 & 52.02 & 0.047 & \textbf{50.63} & \textbf{0.006}  & 51.51 & 0.054 & \textbf{50.27} & \textbf{0.006}  & 64.20 & 0.033 & \textbf{54.02} & \textbf{0.006}  & \textbf{51.23} & 0.034 & 52.79 & \textbf{0.007} \\ \hline
L-95  & Q-1 & 51.97 & 0.215 & \textbf{48.66} & \textbf{0.119}  & 52.85 & 0.240 & \textbf{48.02} & \textbf{0.119}  & 51.33 & 0.244 & \textbf{48.23} & \textbf{0.119}  & 53.61 & 0.182 & \textbf{49.47} & \textbf{0.119} \\ \hline
L-95  & B-1 & 50.89 & 0.031 & \textbf{47.48} & \textbf{0.010}  & 51.06 & 0.034 & \textbf{46.12} & \textbf{0.010}  & 50.70 & 0.034 & \textbf{46.67} & \textbf{0.010}  & 52.15 & 0.023 & \textbf{49.05} & \textbf{0.010} \\ \hline
L-95  & L-2 & 33.73 & 0.030 & \textbf{33.42} & \textbf{0.009}  & 34.64 & 0.032 & \textbf{33.45} & \textbf{0.009}  & 41.21 & 0.035 & \textbf{34.91} & \textbf{0.009}  & 34.72 & 0.024 & \textbf{34.22} & \textbf{0.009} \\ \hline
L-95  & Q-2 & 31.71 & 0.081 & \textbf{30.66} & \textbf{0.056}  & 32.30 & 0.095 & \textbf{30.67} & \textbf{0.056}  & 31.99 & 0.095 & \textbf{30.68} & \textbf{0.056}  & 32.52 & 0.070 & \textbf{31.22} & \textbf{0.056} \\ \hline
L-95  & B-2 & 31.22 & 0.021 & \textbf{30.60} & \textbf{0.010}  & 31.80 & 0.023 & \textbf{30.18} & \textbf{0.010}  & 31.41 & 0.021 & \textbf{30.29} & \textbf{0.010}  & 32.27 & 0.019 & \textbf{31.35} & \textbf{0.010} \\ \hline
L-127 & Q-1 & 44.82 & 0.205 & \textbf{42.07} & \textbf{0.119}  & 45.48 & 0.234 & \textbf{41.72} & \textbf{0.119}  & 45.38 & 0.232 & \textbf{41.45} & \textbf{0.119}  & 46.21 & 0.175 & \textbf{43.11} & \textbf{0.119} \\ \hline
L-127 & B-1 & 44.08 & 0.025 & \textbf{41.64} & \textbf{0.010}  & 44.51 & 0.029 & \textbf{40.60} & \textbf{0.010}  & 43.35 & 0.027 & \textbf{40.74} & \textbf{0.010}  & 45.07 & 0.019 & \textbf{42.93} & \textbf{0.010} \\ \hline
L-127 & Q-2 & 27.48 & 0.076 & \textbf{26.59} & \textbf{0.056}  & 27.96 & 0.090 & \textbf{26.52} & \textbf{0.056}  & 27.62 & 0.089 & \textbf{26.49} & \textbf{0.056}  & 28.23 & 0.066 & \textbf{27.18} & \textbf{0.056} \\ \hline
L-127 & B-2 & 26.93 & 0.018 & \textbf{26.71} & \textbf{0.010}  & 27.42 & 0.020 & \textbf{26.13} & \textbf{0.010}  & 27.03 & 0.018 & \textbf{26.09} & \textbf{0.010}  & 28.02 & 0.017 & \textbf{27.28} & \textbf{0.010} \\ \hline
L-206 & B-1 & 36.04 & 0.033 & \textbf{33.43} & \textbf{0.010}  & 36.76 & 0.037 & \textbf{33.20} & \textbf{0.010}  & 36.03 & 0.030 & \textbf{33.00} & \textbf{0.010}  & 36.46 & 0.022 & \textbf{34.67} & \textbf{0.010} \\ \hline
L-206 & B-2 & 22.33 & 0.022 & \textbf{21.88} & \textbf{0.010}  & 22.73 & 0.025 & \textbf{21.77} & \textbf{0.010}  & 22.25 & 0.020 & \textbf{21.49} & \textbf{0.010}  & 22.72 & 0.018 & \textbf{22.16} & \textbf{0.010} \\ \hline
L-332 & B-1 &    -  &   -  &    -  &   -   & 30.91 & 0.046 & \textbf{28.77} & \textbf{0.010}  &    -  &   -  &    -  &   -   &    -  &   -  &    -  &   -  \\ \hline
L-610 & B-1 &    -  &   -  &    -  &   -   & 25.89 & 0.053 & \textbf{24.49} & \textbf{0.010}  &    -  &   -  &    -  &   -   &    -  &   -  &    -  &   -  \\ \hline
G-94  & L-1 & 65.63 & 0.043 & \textbf{60.03} & \textbf{0.008}  & 70.52 & 0.032 & \textbf{61.65} & \textbf{0.010}  & 95.35 & 0.028 & \textbf{74.04} & \textbf{0.008}  & 64.39 & 0.027 & \textbf{61.46} & \textbf{0.009} \\ \hline
G-94  & Q-1 & 62.24 & 0.204 & \textbf{57.87} & \textbf{0.119}  & 61.64 & 0.238 & \textbf{59.63} & \textbf{0.119}  & 61.69 & 0.240 & \textbf{54.17} & \textbf{0.119}  & 58.75 & 0.171 & \textbf{55.93} & \textbf{0.119} \\ \hline
G-94  & B-1 & 59.26 & 0.033 & \textbf{55.71} & \textbf{0.010}  & 60.70 & 0.033 & \textbf{57.10} & \textbf{0.010}  & 60.66 & 0.033 & \textbf{57.44} & \textbf{0.010}  & 57.18 & 0.023 & \textbf{54.13} & \textbf{0.010} \\ \hline
G-94  & L-2 & 35.63 & 0.033 & \textbf{34.73} & \textbf{0.009}  & 39.09 & 0.031 & \textbf{36.69} & \textbf{0.009}  & 50.20 & 0.034 & \textbf{38.79} & \textbf{0.009}  & 36.39 & 0.025 & \textbf{35.02} & \textbf{0.009} \\ \hline
G-94  & Q-2 & 34.37 & 0.081 & \textbf{33.56} & \textbf{0.056}  & 34.70 & 0.094 & \textbf{34.32} & \textbf{0.056}  & 34.10 & 0.095 & \textbf{32.71} & \textbf{0.056}  & 34.74 & 0.068 & \textbf{33.53} & \textbf{0.056} \\ \hline
G-94  & B-2 & 34.65 & 0.023 & \textbf{33.82} & \textbf{0.010}  & 35.12 & 0.023 & \textbf{34.11} & \textbf{0.010}  & 33.20 & 0.022 & \textbf{32.17} & \textbf{0.010}  & 34.19 & 0.020 & \textbf{33.46} & \textbf{0.010} \\ \hline
G-127 & Q-1 & 49.13 & 0.202 & \textbf{47.71} & \textbf{0.119}  & 52.00 & 0.230 & \textbf{48.45} & \textbf{0.119}  & 48.57 & 0.230 & \textbf{47.66} & \textbf{0.119}  & 49.37 & 0.168 & \textbf{46.52} & \textbf{0.119} \\ \hline
G-127 & B-1 & 48.82 & 0.026 & \textbf{46.04} & \textbf{0.010}  & 50.39 & 0.027 & \textbf{46.65} & \textbf{0.010}  & 47.76 & 0.026 & \textbf{46.47} & \textbf{0.010}  & 47.77 & 0.021 & \textbf{45.31} & \textbf{0.010} \\ \hline
G-127 & Q-2 & 29.29 & 0.076 & \textbf{28.40} & \textbf{0.056}  & 29.94 & 0.089 & \textbf{28.66} & \textbf{0.056}  & 29.02 & 0.088 & \textbf{28.10} & \textbf{0.056}  & 29.83 & 0.064 & \textbf{28.57} & \textbf{0.056} \\ \hline
G-127 & B-2 & 28.90 & 0.019 & \textbf{28.59} & \textbf{0.010}  & 29.47 & 0.020 & \textbf{28.57} & \textbf{0.010}  & 28.23 & 0.018 & \textbf{27.42} & \textbf{0.010}  & 29.54 & 0.018 & \textbf{28.72} & \textbf{0.010} \\ \hline
G-204 & B-1 & 39.83 & 0.035 & \textbf{37.41} & \textbf{0.010}  & 41.39 & 0.036 & \textbf{38.52} & \textbf{0.010}  & 40.06 & 0.030 & \textbf{36.48} & \textbf{0.010}  & 38.84 & 0.025 & \textbf{36.99} & \textbf{0.010} \\ \hline
G-204 & B-2 & 24.13 & 0.023 & \textbf{23.43} & \textbf{0.010}  & 24.81 & 0.025 & \textbf{24.02} & \textbf{0.010}  & 23.80 & 0.021 & \textbf{23.01} & \textbf{0.010}  & 24.14 & 0.020 & \textbf{23.34} & \textbf{0.010} \\ \hline
G-325 & B-1 &    -  &   -  &    -  &   -   & 35.46 & 0.047 & \textbf{32.74} & \textbf{0.010}  &    -  &   -  &    -  &   -   &    -  &   -  &    -  &   -  \\ \hline
\end{tabular}%
}
\caption{Evaluation Perplexity of LLaMA and GPT models pre-trained on C4 with hypersphere optimizers, with and without CEWT. The first two columns are ``Model'' (L for LLaMA, G for GPT, with a dash followed by the number of parameters in millions) and ``Q-B'' which stands for Quantizer-Bits (L for LSQ, Q for QuEST, B for BBQ, with a dash followed by the \textbf{activation and weight precision} in number of bits). PPL stands for perplexity, and RBM stands for rounding boundary mass (Equation~\ref{eqn:rbm}) of pre-round weights (Algorithms \ref{alg:bbq}, \ref{alg:quest}, and \ref{alg:lsq}). }
\label{tab:main}
\end{table}

\section{Limitations}
CEWT relies on the law of large numbers to guarantee that a random permutation of weights look indistinguishable from Gaussian noise. Therefore, for extremely small models with only a few hundred parameters in a single layer, CEWT does not work well as it forbids making small changes to parameters. Our ablation studies in Section~\ref{sec:sweep_dniter} shows that CEWT generally works well as long as the number of parameters in a layer exceed $2^{10}$.

CEWT improves accuracy/reduces loss by suppressing rounding boundary weight oscillation, which is most severe during Quantization-Aware \textbf{Pre-Training} (QAPT), where dataset size is large and training time is long. Section~\ref{sec:sweep_dniter} shows that when training time is short and dataset size is much smaller compared to model size, weight oscillation is less severe, and the accuracy improvement of CEWT can be small and inconsistent. These insights from Section~\ref{sec:sweep_dniter} suggest CEWT might not be suitable for QAFT. However, Section~\ref{sec:eval_deep} shows that CEWT consistently improves accuracy in large-scale QAPT.

CEWT maintains the zero-mean Gaussian constraint using argsort, which has an $O(mn\log(mn))$ time complexity. The runtime overhead can become more severe for wider models, but the overhead can potentially be reduced with engineering tweaks. For example, CEWT may be applied to each channel individually, reducing the time complexity to $O(mn\log(n))$. In practice we observe CEWT results in a geomean increase of 4\% in training time, while matching the accuracy of existing methods with the best hyperparameters found using \textbf{computationally expensive grid searches}.

\section{Conclusion}
In this work, we introduce optimization with Constrained Empirical Weight disTributions (CEWT), an optimizer post-update step that projects weights to the nearest point in weight space whose empirical distribution matches a zero-mean Gaussian. CEWT enforces the zero-mean Gaussian assumption, which is built into the design of many existing quantizers. CEWT suppress the irregularly high density spikes that tend to occur due to the weight oscillation problem, thereby reducing gradient noise and improving accuracy.

\section*{AI use statement}



We used AI tools to paraphrase less than five sentences in this paper. We use AI tools to review our paper and manually address the reviews. We used in-IDE auto-complete features of AI tools for code writing. We used AI coding agents to review our code and check for bugs, and we manually addressed the reviews. We occasionally accept coding-agent-proposed code changes that are less than 10 lines after carefully reviewing. We take responsibility for the final content of this work.

\bibliography{iclr2027_conference}
\bibliographystyle{iclr2027_conference}

\newpage
\appendix

\clearpage
\begingroup
\raggedbottom
\section{Examples of Rounding Boundary Weight Oscillation}\label{sec:oscillation_example}
In this section, we discuss two example problems in which rounding boundary weight oscillation occurs. The first example is a one-dimensional optimization problem that reveals some insight on the root cause of rounding boundary weight oscillation. The second example is a two-dimensional optimization problem that shows the oscillation of one quantized parameter at one of its rounding boundaries can introduce significant variance to the partial derivative of another parameter due to parameter interdependency.
\subsection{A 1D Example}\label{sec:exp_1d}
\begin{figure}[!htbp]
    \centering
    \includegraphics[width=\linewidth]{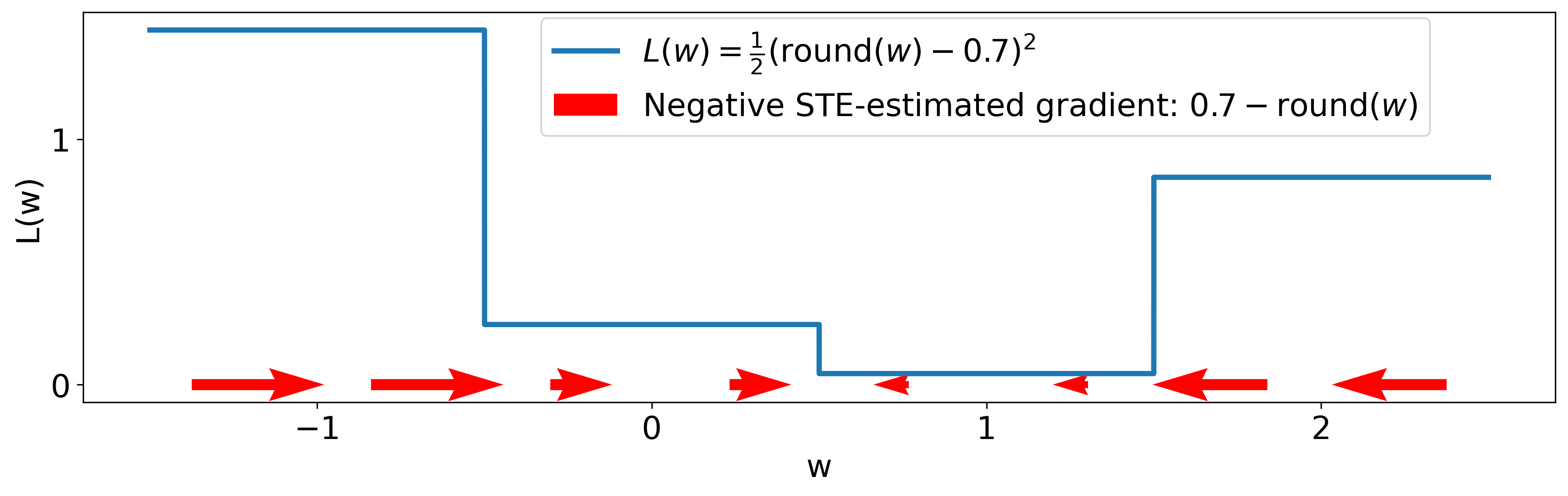}
    \caption{Visualization of $L(w)=\frac{1}{2}(\text{round}(w)-0.7)^2$ as a function and its negative STE-estimated gradient $0.7 - \text{round}(w)$ as a one-dimensional vector field. The direction of an arrow represents its sign (left for negative and right for positive), and the size of an arrow represents its magnitude. }
    \label{fig:exp_1d}
\end{figure}

The root cause of rounding boundary weight oscillation can be explained with a simple example.
Consider the problem of finding the closest integer to $0.7$. To solve this problem with gradient descent, we reformulate it to the following alternative problem which has a continuous optimization variable $w \in \mathbb{R}$:
\begin{equation}\label{eqn:example_problem}
    \begin{split}
        \underset{w \in \mathbb{R}}{\text{min. }} & \frac{1}{2}(q - 0.7)^2\\
        &q = \text{round}(w)
    \end{split}
\end{equation}
Once a solution of this alternative problem is found, denoted as $w^*$, we can calculate $\text{round}(w^*)$ which is the solution to the original problem of finding the closest integer to 0.7. Problem~\ref{eqn:example_problem} has more than one solution. Any $w'$ that satisfies $0.5 \le w' < 1.5$ is a valid solution as it suggests that the closest integer to 0.7 is $\text{round}(w') = 1$.

Let $L(w) = \frac{1}{2}(q - 0.7)^2$. We invoke the chain rule to attempt to calculate ${dL / dw}$. However, the round function is not differentiable, and we use the Straight-Through Estimator (STE) to estimate the derivative of the round function as $1$, arriving at the following approximated gradient:
\begin{equation}
    \frac{dL}{dw} = \frac{dL}{dq} \cdot \frac{dq}{dw} = (q - 0.7) \cdot \text{round}'(w) \underset{\text{STE}}{\approx} (q - 0.7) \cdot 1= \text{round}(w) - 0.7
\end{equation}
With gradient descent, our update rule is therefore:
\begin{equation}
    w \leftarrow w - \eta (\text{round}(w) - 0.7)
\end{equation}
where $\eta$ is a sufficiently small learning rate. 
Figure~\ref{fig:exp_1d} visualizes $L(w)$ and its negative STE estimated gradient $0.7 - \text{round}(w)$ as a one-dimensional vector field. Figure~\ref{fig:exp_1d} shows when $L(w)$ is \textit{not} minimized, the negative STE-estimated gradient points towards the minimum solution. However, when $L(w)$ \textit{is} minimized (i.e. $0.5\le w < 1.5$), the negative STE-estimated gradient evaluates to $0.7 - 1 = -0.3$, which is a negative number and \textit{not} zero. This means even though an optimal solution has been found, $w$ will continue to decrease. Eventually, $w$ becomes slightly smaller than the rounding boundary $0.5$, in which case 
the negative STE-estimated gradient evaluates to $0.7 - 0 = 0.7$, a positive number. This means $w$ will increase until $w$ becomes larger than $0.5$ again. In other words, $w$ will continuously oscillate between the left and right side of $0.5$ and never stop moving, and $q=\text{round}(w)$ will constantly oscillate between $0$ and $1$. 
This oscillation happens because $q$ is not a continuous variable, and therefore $q$ cannot encode $q=0.7$, the stationary point that zeros the estimated gradient $q - 0.7$. 

\subsection{A 2D Example}\label{sec:exp_2d}
\begin{figure}[!htbp]
    \centering
    \includegraphics[width=\linewidth]{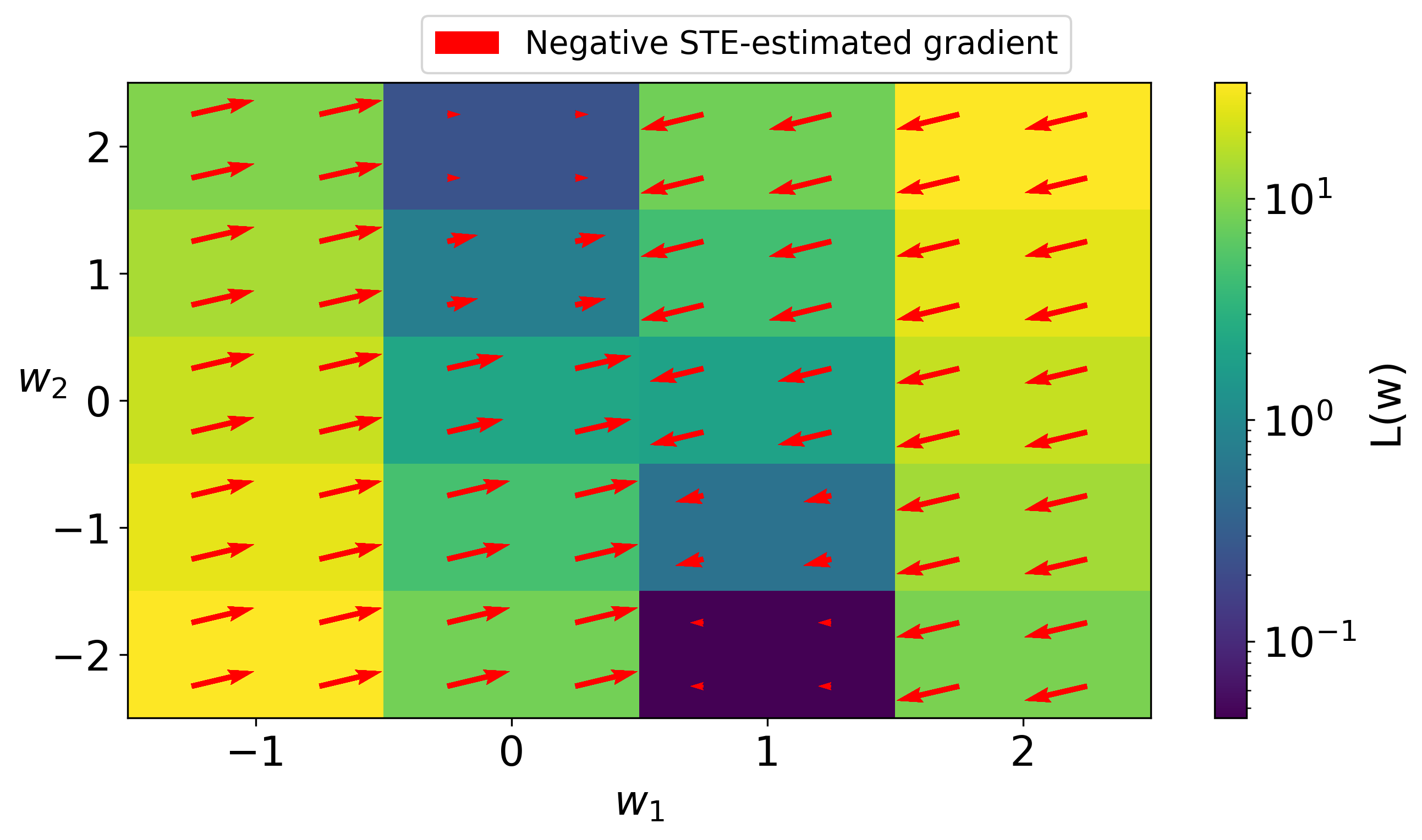}
    \caption{Visualization of $L(w)=\frac{1}{2}(q_1 - 0.7)^2 + \frac{1}{2}(4q_1 - q_2-2)^2$ as a function and its negative STE-estimated gradient $\left[-(q_1 - 0.7)-4\cdot(4q_1 + q_2-2), -(4q_1 + q_2-2)\right]^\top$ as a two-dimensional vector field. Note that $q_1 = \text{round}(w_1)$ and $q_2 = \text{round}(w_2)$. Darker colored regions have lower loss than brighter regions.}
    \label{fig:exp_2d}
\end{figure}
While rounding boundary weight oscillation appears harmless, we show it can significantly increase gradient variance in high-dimensional problems due to parameter interdependency. Consider the following 2D problem:
\begin{equation}\label{eqn:example_problem}
    \begin{split}
        \underset{w \in \mathbb{R}^2}{\text{min. }} & \frac{1}{2}(q_1 - 0.7)^2 + \frac{1}{2}(4q_1 - q_2-2)^2\\
        \text{s.t. } & q_1 = \text{round}(w_1) \\
        & q_2 = \text{round}(w_2)
    \end{split}
\end{equation}
Let $L(w) = \frac{1}{2}(q_1 - 0.7)^2 + \frac{1}{2}(4q_1 - q_2-2)^2$. We invoke the chain rule to attempt to calculate the partial derivatives with respect to $w_1$ and $w_2$, but the round function is not differentiable. So, we use the Straight-Through Estimator (STE) to estimate the derivative of the round function as 1, arriving at the following expressions:
\begin{equation}\label{eqn:exp_2d_pd}
\begin{split}
    \frac{\partial L}{\partial w_1} &= \frac{\partial L}{\partial q_1} \cdot \frac{\partial q_1}{\partial w_1} \underset{\text{STE}}{\approx} (q_1 - 0.7)+4\cdot(4q_1 + q_2-2) \\
    \frac{\partial L}{\partial w_2} &= \frac{\partial L}{\partial q_2} \cdot \frac{\partial q_2}{\partial w_2} \underset{\text{STE}}{\approx} (4q_1 + q_2-2)\\
\end{split}
\end{equation}
Therefore, we use the following STE-estimate gradient:
\begin{equation}
    \nabla w = \left[\frac{\partial L}{\partial w_1}, \frac{\partial L}{\partial w_2}\right]^\top \underset{\text{STE}}{\approx} \begin{bmatrix}
        (q_1 - 0.7)+4\cdot(4q_1 + q_2-2)\\
        (4q_1 + q_2-2)
    \end{bmatrix}
\end{equation}

Figure~\ref{fig:exp_2d} visualizes $L(w)$ and its negative STE-estimated gradient. Figure~\ref{fig:exp_2d} shows that when $w_1$ and $w_2$ are initialized close to zero, the negative estimated gradient has a strong tendency to immediately push $w_1$ towards the line $w_1 = 0.5$, where $w_1$ will start to oscillate (as long as $-1.5 \le w_2 < 1.5$). When $w_1 \approx 0.5$, the loss can no longer be significantly reduced by changing $w_1$, but changing $w_2$ (by pushing it upward towards the line $w_2 = 2$ or by pushing it downwards towards $w_2 = -2$) can reduce the loss significantly. However, changing $w_2$ can be challenging when $w_1\approx 0.5$. When $w_1$ is close to, but less than $0.5$, the negative estimated gradient has a tendency to increase $w_2$. When $w_1$ is close to, but greater than or equal to $0.5$, the negative estimated gradient has a tendency to decrease $w_2$. Since $w_1$ constantly oscillates around $w_1=0.5$, the trajectory of $w_2$ will be noisy, with some iterations canceling the progress of previous iterations. The core reason is $\frac{\partial L}{\partial w_2}$ \textbf{depends} (see Equation~\ref{eqn:exp_2d_pd}) on the value of $q_1$. When $w_1$ oscillates around $w_1=0.5$, the value of $q_1$ jumps between $0$ and $1$, thereby increasing the variance of  $\frac{\partial L}{\partial w_2}$. To summarize, \textbf{due to parameter interdependency, rounding boundary weight oscillation can introduce significant gradient variance in high-dimensional problems}.

\section{Is $R$ a hyperparameter of CEWT?}\label{sec:sweep_r}
In this section, we discuss that $R$ is \textbf{not} a hyperparameter of CEWT or HO, as \textbf{every} deep learning pre-training script choose a value of $R$ by randomly initializing weight matrix $W\in \mathbb{R}^{m\times n}$ from a zero-mean distribution with a variance of $\sigma^2$. For a sufficiently large $mn$, the Frobenius norm radius is $R \approx \sigma\sqrt{mn}$ and the spectral norm radius is $R \approx \sigma(\sqrt{m} + \sqrt{n})$. Therefore, at random initialization time, every deep learning model lie on some radius-$R$ hypersphere. Hypersphere optimizers (e.g. AdamH/MuonH) proposes to \textbf{keep} the model on this radius-$R$ hypersphere, thereby \textbf{reusing} the hyperparameter $R$, bounding the Frobenius norm of $W$, and removing the need for weight decay. In this work, we use the Lecun initialization~\citep{lecun-98x} which results in the Frobenius norm radius $R\approx \sqrt{m}$. We additionally perform ablation studies on other possible weight initialization strategies/values of R: PyTorch~\citep{NEURIPS2019_bdbca288} default initialization, Spectral $\mu P$\citep{yang2024spectralconditionfeaturelearning}, and the GPT-2~\citep{Radford2019LanguageMA} initialization strategy. Table~\ref{tab:sweep_r} shows that \textbf{the perplexity reduction of CEWT compared to HO is consistent across multiple values of $R$}. For clarity, Table~\ref{tab:hp_exp} lists multiple combinations of regular/hypersphere optimizers and weight oscillation suppression methods and shows that CEWT has the \textbf{least} number (one to be exact) of hyperparameters among all combinations. In particular, Table~\ref{tab:hp_exp} shows $R$ is \textbf{not} a hyperparameter \textbf{introduced by CEWT} but is a part of \textbf{every} deep learning pre-training script that performs weight random initialization. 
\begin{table}[H]
\centering
\resizebox{\textwidth}{!}{
\renewcommand{\arraystretch}{2.5} 
\begin{tabular}{|l|c|c|c|c|c|}
\hline
Name      & Lecun                     & PyTorch                           & Spectral $\mu P$                  & GPT-2 (non-residual) & GPT-2 (residual) \\ \hline
Init Distribution & Normal                    & Uniform                           & Normal                       & Normal               & Normal           \\ \hline
Init Mean  & $0$ & $0$ & $0$ & $0$                 & $0$    \\ \hline
Init Std  & $\sqrt{\frac{1}{n}}$ & $\sqrt{\frac{1}{3n}}$ & $\frac{\sqrt{m/n}}{\sqrt{m} + \sqrt{n}}$ & $0.02$                 & $\frac{0.02}{\sqrt{2l}}$    \\ \hline
Frobenius norm $R$  &  $\sqrt{m}$                     &  $\sqrt{m/3}$                         &  $\frac{m}{\sqrt{m} + \sqrt{n}}$          & $0.02\sqrt{mn}$          & $0.02\sqrt{\frac{mn}{2l}}$            \\ \hline
AdamH Perp & 31.78 & 31.52 & 31.39 & \multicolumn{2}{c|}{30.91} \\ 
Adam-CEWT Perp & \textbf{30.12} & \textbf{30.22} & \textbf{29.82} & \multicolumn{2}{c|}{\textbf{29.77}} \\ \hline
\end{tabular}%
}
\caption{Robustness of CEWT over different values of $R$. $m$ is the number of output features of a linear layer. $n$ is the number of input features of a linear layer. $l$ is the number of layers in a model. Experiments are conducted using a LLaMA-95M with weights and activations quantized to 2-bit using BBQ. The model is pre-trained on 3 billion C4 tokens. We report perplexity (Perp).}
\label{tab:sweep_r}
\end{table}

\begin{table}[H]
\centering
\resizebox{\textwidth}{!}{%
\renewcommand{\arraystretch}{1.3} 
\begin{tabular}{|l|c|c|c|c|}
\hline
                               & Weight Init HP (R) & Weight Decay HP & RB Weight Oscillation HP & \# HPs     \\ \hline
Adam                           & Yes                & Yes             & No                       & 2          \\ \hline
AdamH                          & Yes                & No              & No                       & \textbf{1} \\ \hline
Adam + Periodic Interpolation  & Yes                & Yes             & Yes                      & 3          \\ \hline
AdamH + Periodic Interpolation & Yes                & No              & Yes                      & 2          \\ \hline
Adam + Dampening Loss          & Yes                & Yes             & Yes                      & 3          \\ \hline
AdamH + Dampening Loss         & Yes                & No              & Yes                      & 2          \\ \hline
Adam + PQN                     & Yes                & Yes             & No                       & 2          \\ \hline
AdamH + PQN                    & Yes                & No              & No                       & \textbf{1} \\ \hline
Adam-CEWT                      & Yes                & No              & No                       & \textbf{1} \\ \hline
\end{tabular}%
}
\caption{The number of hyperparameters among weight initialization radius $R$, weight decay, and rounding boundary weight oscillation, for multiple combinations of regular/hypersphere optimizers and oscillation suppression methods. A ``yes'' means the hyperparameter is used. A ``no'' means the hyperparameter is \textit{not} used.}
\label{tab:hp_exp}
\end{table}

\section{Metrics To Quantify Rounding Boundary Weight Oscillation}\label{sec:quantify_wo}
\begin{figure}[H]
    \centering
    \includegraphics[width=\linewidth]{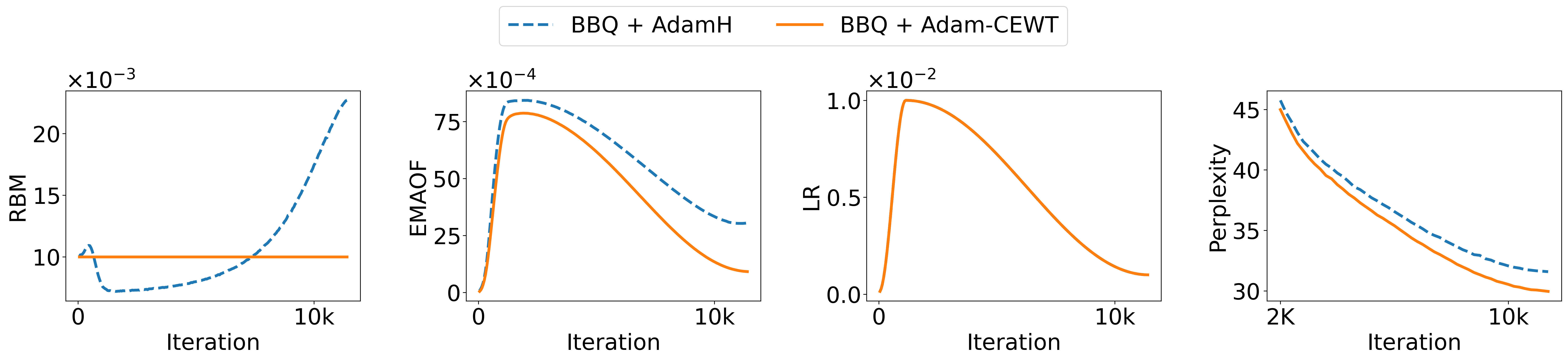}
    \includegraphics[width=\linewidth]{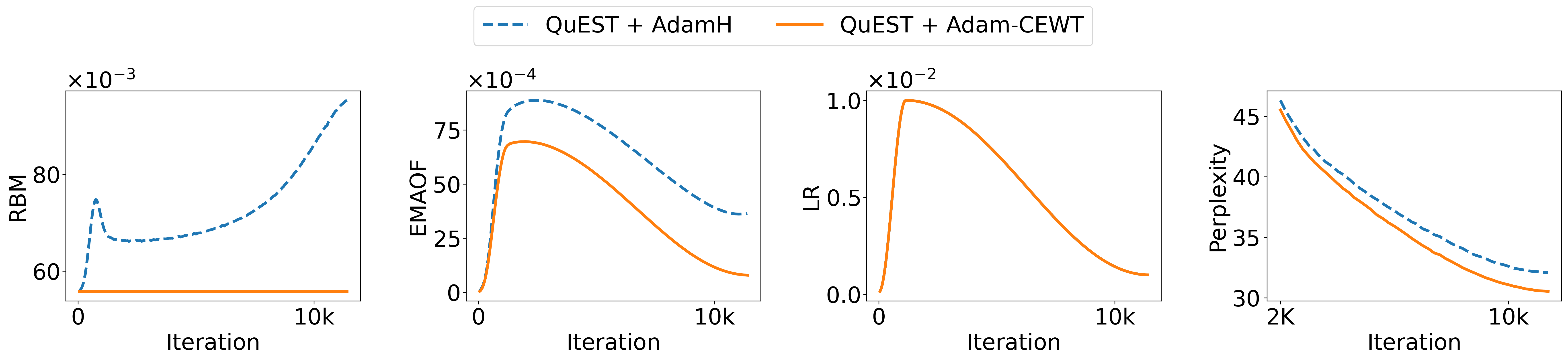}
    \caption{Rounding Boundary Mass (RBM), EMA Oscillation Frequency (EMAOF), learning rate (LR), and perplexity of a LLaMA-95M model with 2-bit weights and activations, pretrained on 3B C4 tokens. EMAOF is a more direct indicator of rounding boundary weight oscillation, but it correlates with LR and is not scale-invariant. RBM is scale-invariant, and RBM shows CEWT regulates the number of parameters close to rounding boundary \textit{by design}. }
    \label{fig:ema_osc_freq_bbq}
\end{figure}
In this work, we use two metrics to quantify weight oscillation: EMA Oscillation Frequency~\citep{pmlr-v162-nagel22a} and Rounding Boundary Mass (RBM) which is inspired by \citet{10.5555/3618408.3619311}. EMA Oscillation Frequency (EMAOF) quantifies the frequency of rounding boundary weight oscillation, but calculating EMAOF introduces training storage overhead, and EMAOF correlates with the learning rate. RBM quantifies the amount of weights sufficiently close to a rounding boundary, has no training storage overhead, and is invariant of learning rate.

EMA Oscillation Frequency, denoted $f^{(t)}$, is calculated as follows. Denoted $m \in (0, 1)$ as the momentum of the exponential moving average. Denote $w_{\text{int}}^{(t)}$ as the integer value, or quantization bin number, of a parameter $w$ at training iteration $t$. Denote $\Delta^{(t)}\in \mathbb{Z}$ as a storage variable initialized to zero.  Note $f^{(t)}$ is also a storage variable initialized to zero. Variables $\Delta^{(t)}$ and $f^{(t)}$ are updated every training iteration as follows:
\begin{equation}
\begin{split}
    \Delta_{\text{int}} &= w_{\text{int}}^{(t)} - w_{\text{int}}^{(t-1)} \\
    o &= \mathbf{1}_{\left(\text{sign}(\Delta_{\text{int}}\right) \neq \text{sign}(\Delta_{\text{int}}^{\tau})) \land \left(\Delta_{int} \neq 0 \right)} \\
    f^{(t)} &= m \cdot o + (1 - m) \cdot f^{(t-1)} \\
    \Delta^{(t)} &= \begin{cases}
        \Delta_{\text{int}}, &\text{ if } o = 1\\
        \Delta^{(t-1)}, &\text{ otherwise}
    \end{cases}
\end{split}
\end{equation}
where $\mathbf{1}_{\text{cond}}$ returns one if cond is true and a zero if cond is false. We use $m=0.1$ for our toy problem evaluation in Tables~\ref{tab:toy} and \ref{tab:toy_grad_est}. We use $m=0.01$ for our deep learning experiments in Figure~\ref{fig:ema_osc_freq_bbq}. For an experiment with a total of $T$ training iterations, we report $f^{(T)}$ averaged across all parameters in the model. EMAOF lies in the range $[0, 1]$. $f^{(t)}=1$ indicates that $w$ crosses a rounding boundary every iteration, and that the direction in which $w$ crosses that rounding boundary flips every iteration. $f^{(t)}=0$ indicates that there is no rounding boundary weight oscillation recently. EMAOF is a direct indicator of rounding boundary weight oscillation, but it introduces a significant training overhead of two storage variables per parameter: $\Delta^{(t)}$ and $f^{(t)}$. Additionally, Figure~\ref{fig:ema_osc_freq_bbq} shows EMAOF correlates with learning rate and is therefore not scale-invariant during training and across different learning rates or learning rate schedules. For example, a model may have extremely sharp density spikes near rounding boundaries, but if the learning rate is 0, oscillation \textit{cannot} happen and therefore EMAOF is 0. Therefore, for large scale-experiments (e.g. Table~\ref{tab:main}), we prefer and use RBM as it introduces no training storage overhead and is scale-invariant.  

Rounding Boundary Mass (RBM) quantifies the amount of weights which are sufficiently close to their closest rounding boundary. Given a list of pre-rounding\footnote{The definition of pre-rounding weights varies across quantizers. See Algorithms~\ref{alg:bbq}, \ref{alg:quest}, and \ref{alg:lsq} for examples.} weights $w\in \mathbb{R}^n$, RBM$:\mathbb{R}^n\rightarrow[0,1]$ is defined as: 
\begin{equation}\label{eqn:rbm}
    \text{RBM}(w)=\frac{1}{n} \sum_{i=0}^{n-1} \mathbf{1}_{|w_i-(\text{floor}(w_i) + 0.5)| < \epsilon}
\end{equation}
where $\epsilon$ is chosen to be 0.005 following \citet{10.5555/3618408.3619311}, and $\mathbf{1}_{\text{cond}}$ returns one if cond is true and a zero if cond is false. RBM is a estimation to the area under the rounding boundary spikes, and therefore also functions as a proxy to the severity of weight oscillation. 
\textbf{RBM does not introduce any training overhead}, and therefore we prefer RBM for large-scale experiments.

\section{Details of Evaluation on Toy Problem}\label{sec:toy_details}
In this section we discuss details of the rounding boundary weight oscillation suppression algorithms we used in Section~\ref{sec:eval_toy}. We reiterate Problem~\ref{eqn:toy} here:
\begin{equation}\label{eqn:toy_problem_expanded}
    \begin{split}
        \underset{W}{\text{min}} \text{ } &\frac{1}{2} ||XQ - XW^*||_F^2 
        \\
        \text{ s.t. } &Q = s\cdot \lfloor 1 / s \cdot W \rceil\\
        &||W||_F = ||W^*||_F
    \end{split}
\end{equation}
where $s\in \mathbb{R}$, $X \in \mathbb{R}^{n\times d}$, $W, W^* \in \mathbb{R}^{d\times d}$, and $n = d = 2^{10}$. Elements of $X$ are i.i.d. sampled from $N(0, 1)$; elements of $W$ are initialized from $N(0, 1/d)$; elements of $W^*$ are i.i.d. sampled from $N(0, 1/d)$; and $s$ is chosen to be $1/\sqrt{d}$. Note that $X$ and $W^*$ are constants (already sampled and fixed), whereas $W$ is randomly initialized and learnable. Note that Problem~\ref{eqn:example_problem} is an instance of Problem~\ref{eqn:toy_problem_expanded} with $n=d=1$, $s=1$, $X=[1]$, and $W^*= [0.7]$. We use the AdamH optimizer, a learning rate of 0.1, and the Cosine Annealing scheduler. We train a total of 200 iterations. 

\subsection{Why this Problem?}\label{sec:toy_expl}
While Problem~\ref{eqn:toy_problem_expanded} (or Problem \ref{eqn:toy}) is simple, it simulates multiple phenomena observed in QAT of deep networks. First, Problem~\ref{eqn:toy_problem_expanded} recreates weight oscillation. Second, weight oscillation introduces detrimental noise in the optimization process because of \textbf{parameter interdependency}. Specifically, the partial derivative with respect to the quantized parameter $Q_{i,j}$ is:
\begin{equation}
    \frac{\partial L}{\partial Q_{i,j}} = [X^T(XQ-XW^*)]_{i,j} = \sum_{k=0}^{d-1} [X^TX]_{i,k} \cdot (Q_{k,j} - W^*_{k,j})
\end{equation}
In other words, how $Q_{i,j}$ should move \textbf{depends} on the value of $Q_{0, j}$, $Q_{1, j}$, $\cdots$, and $Q_{(d-1), j}$. If any of these parameters starts oscillating (constantly switches between two quantization bins), noise is introduced into the movement of $Q_{i,j}$. Third, the solution to Problem~\ref{eqn:toy_problem_expanded} need not be unique, which simulates the phenomenon that there can be many approximate solutions
(local minima) in the loss landscapes of deeper networks. For example, the loss in Problem~\ref{eqn:toy_problem_expanded} is zeroed when $Q=W^*+E$ as long as $XE=0$, which means there can be multiple non-zero options for $E$ when the null space of $X$ is non-trivial. 
If $W^*$ cannot be perfectly represented/encoded by $Q$ due to quantization constraint, the optimization process may find some $W^*+E$ (where $E\not\approx 0$) that can be better represented by $Q$. 
Fourth, the constraint $||W||_F = ||W^*||_F$ simulates the use of weight decay or hypersphere optimization, in that when multiple solutions exist, ones with reasonably small Frobenius norms are preferred.

\subsection{STE}
When we use the STE~\citep{bengio2013estimatingpropagatinggradientsstochastic}, we estimate the gradient as the following:
\begin{equation}\label{eqn:toy_expand_ste}
    G = X^T(XQ - XW^*)
\end{equation}
The estimated gradient is further processed by AdamH to produced a weight update. AdamH also maintains the Frobenius norm constraint.

\subsection{PQN}\label{sec:pqn_exp}
When we use PQN~\citep{Shin_2023_CVPR}, the optimization objective changes to the following:
\begin{equation}\label{eqn:pqn_toy_objective}
    \begin{split}
        \underset{W}{\text{min}} \text{ } & E_{\Xi_{i, j}\sim U(-\frac{1}{2}, \frac{1}{2}), \forall i, j}\left[\frac{1}{2} ||XQ' - XW^*||_F^2 \right]
        \\
        \text{ s.t. } &Q' = s\cdot ( 1 / s \cdot W + \Xi) = W + s\Xi\\
        &||W||_F = ||W^*||_F
    \end{split}
\end{equation}
In other words, the rounding error is simulated with addition of a noise matrix $\Xi \in \mathbb{R}^{m \times n}$ whose elements are i.i.d. sampled from $U(-\frac{1}{2}, \frac{1}{2})$. During training, we perform Monte Carlo sampling:
\begin{equation}
    \begin{split}
        \Xi_{i, j} &\sim U(-\frac{1}{2}, \frac{1}{2}), \forall i, j \in \{0, 1, \cdots, d-1\} \\
        L(W, \Xi) &= \frac{1}{2}||X(W + s\Xi) - XW^*||_F^2 \\
        \frac{\partial L(W, \Xi)}{\partial W} &= X^T(X(W + s\Xi) - XW^*)
    \end{split}
\end{equation}
and we use a randomly sampled $\frac{\partial L(W, \Xi)}{\partial W}$ to perform weight update.

To understand the expected behaviour of training, we perform the following derivation on our loss function in Equation~\ref{eqn:pqn_toy_objective}:
\begin{equation}
    \begin{split}
        E\left[\frac{1}{2} ||XQ' - XW^*||_F^2\right]
        &= \frac{1}{2} E[||XQ' - XW^*||_F^2]\\
        &= \frac{1}{2} ||E[XQ' - XW^*]||_F^2 + \frac{1}{2} \sum_{\forall i, j}\text{Var}([XQ' - XW^*]_{i,j}) \\
        &= \frac{1}{2} ||E[X(W + s\Xi - W^*)]||_F^2 + \frac{1}{2} \sum_{\forall i, j}\text{Var}([X(W + s\Xi - W^*)]_{i,j})\\
        &= \frac{1}{2} ||X \cdot E[(W + s\Xi - W^*)]||_F^2 + \frac{1}{2} \sum_{\forall i, j}\text{Var}([sX\Xi]_{i,j}) \\
        &= \frac{1}{2} ||X \cdot (W + s \cdot E[\Xi] - W^*)||_F^2 + \frac{1}{2} \sum_{\forall i, j} \frac{s^2}{12}||X_i||_2^2 \\
        &= \frac{1}{2} ||X \cdot (W - W^*)||_F^2 + C \\
        &= \frac{1}{2} ||XW - XW^*||_F^2 + C \\
        &= \frac{nd}{2} \cdot \text{MSE}(XW, XW^*) + C \\
    \end{split}
\end{equation}
which suggests that the modified loss function is simply $\text{MSE}(XW, XW^*)$ scaled by a constant and shifted by a constant. However, Table~\ref{tab:toy} shows PQN achieves suboptimal MSE$(XQ, XW^*)$ despite excelling in reducing MSE$(XW, XW^*)$. In contrast, the STE and CEWT are capable of finding a $W$ that performs better in terms of the true objective MSE$(XQ, XW^*)$. 
This suggests that Problem~\ref{eqn:pqn_toy_objective} is a biased objective, or that a solution to Problem~\ref{eqn:pqn_toy_objective} is not a solution to Problem~\ref{eqn:toy_problem_expanded}.

\subsection{Dampening Loss (DL)}
When we use DL~\citep{pmlr-v162-nagel22a}, the optimization objective changes to the following:
\begin{equation}\label{eqn:toy_problem_dl}
    \begin{split}
        \underset{W}{\text{min}} \text{ } &\frac{1}{2} ||XQ - XW^*||_F^2 + \frac{\alpha}{2} ||W - Q||_F^2 
        \\
        \text{ s.t. } &Q = s\cdot \lfloor 1 / s \cdot W \rceil\\
        &||W||_F = ||W^*||_F
    \end{split}
\end{equation}
where $\alpha$ is a regularization hyperparameter.
We estimate the gradient as follows:
\begin{equation}
    G = X^T(XQ - XW^*) + \alpha (W - Q)
\end{equation}
Note that the second gradient term $\alpha (W - Q)$ assumes~\citep{pmlr-v162-nagel22a} $Q$ is a constant in the second loss term $||W - Q||_F^2$.

Our findings agree with \citet{pmlr-v162-nagel22a}, in that $\alpha$ should be small at the beginning of training and should be gradually increased during training. An  $\alpha$ that is too large too early generally prevents convergence.

\subsection{Iterative Freezing (IF)}
When we use IF~\citep{pmlr-v162-nagel22a}, our weight update algorithm is Algorithm~\ref{alg:if}. Our findings agree with \citet{pmlr-v162-nagel22a}, in that hyperparameter $f_{th}$ should be large at the beginning of training and gradually decrease during training. An $f_{th}$ that is too small too early prevents convergence.
\begin{algorithm}[H]
\caption{Iterative Freezing applied to Problem~\ref{eqn:toy_problem_expanded}}
\label{alg:if}
\begin{algorithmic}[1]
    \Function{IF}{$f_{th}$, $m$} 
        \State Introduce four variables: $F \in \mathbb{R}^{d\times d}$, $B \in \{\text{False}, \text{True}\}^{d \times d}$, $\Delta \in \mathbb{R}^{d\times d}$, and $W_{\text{EMA}}\in \mathbb{R}^{d\times d}$
        \State $F^{(0)} = 0$
        \State $B^{(0)} = 0$
        \State $\Delta = 0$
        \State $W_{\text{EMA}} = \lfloor 1 / s \cdot W^{(0)}\rceil$
        \For{$t = 1$ to $T$}
            \State $Q = s\cdot \lfloor 1 / s \cdot W^{(t-1)}\rceil$
            \State $G = X^T(XQ - XW^*)$ 
            \State perform optimizer step with gradient $G$ to obtain updated weight $W_{\text{curr}}^{(t)}$, but only update weights at index $(i, j)$ s.t. $B^{(t-1)}_{i,j}$ is True
            \State $\Delta_{\text{int}} = \lfloor 1 / s \cdot W^{(t)}\rceil - \lfloor 1 / s \cdot W^{(t-1)}\rceil$
            \State $O = (\text{sign}(\Delta_{\text{int}}) \neq \text{sign}(\Delta^{(t-1)})) \land (\Delta_{\text{int}} \neq 0)$\Comment{Element-wise ops (sign, $\neq$, and $\land$)}
            \State $F^{(t)} = m \cdot O + (1 - m) \cdot F^{(t-1)}$
            \State $B^{(t)} = B^{(t-1)}  \lor (F^{(t)} > f_{th})$\Comment{Element-wise ops ($\lor$ and $>$)}
            \State $W^{(t)} = \text{where}\left(F^{(t)} > f_{th}, s \cdot W_{\text{EMA}}^{(t-1)}, W_{\text{curr}}\right)$\Comment{``where'' is similar to torch.where}
            \State $W_{\text{EMA}}^{(t)} = m \cdot \lfloor 1 / s \cdot W^{(t-1)}\rceil + (1 - m) \cdot W_{\text{EMA}}^{(t-1)}$
            \State $\Delta^{(t)} = \text{where}\left(O, \Delta_{\text{int}}, \Delta^{(t-1)}\right)$
        \EndFor
    \EndFunction
\end{algorithmic}
\end{algorithm}

\subsection{Confidence Guided Annealing}
When we use CGA~\citep{10.5555/3618408.3619311}, we prune the STE gradient before passing it to the optimizer, as follows:
\begin{equation}
\begin{split}
    D = |(1 / s \cdot W) - (\lfloor 1 / s \cdot W \rfloor + 0.5)| \\
    G = \left(X^T(XQ - XW^*)\right) \odot \underbrace{\mathbf{1}_{D \le \text{BR}_x}}_{\text{pruning mask}}
\end{split}
\end{equation}
where $|\cdot|$, $\lfloor \cdot \rfloor$, $\odot$ are the element-wise absolute value operator, the element-wise floor operator, and the element-wise multiplication operator, respectively; $\mathbf{1}_{\text{cond}}$ returns a one if cond is true or a zero if cond is false; and $\text{BR}_x \in [0, 0.5]$ is a hyperparameter. In practice, we use $0.5 / \text{BR}_x \cdot G$ (instead of $G$) as the actual gradient to compensate for the decrease in gradient norm due to gradient pruning.

We find that $\text{BR}_x$ should be large at the beginning of training and should be gradually decreased during training. A $\text{BR}_x$ that is too small too early prevents convergence.

\subsection{Periodic Interpolation (PI)}
When we use PI~\citep{li2026winqacceleratingquantizationawaretraining}, our weight update algorithm is Algorithm~\ref{alg:pi}. PI introduces hyperparameters $\alpha$ and $K$, both of which need to be tuned. We find $\alpha$ does not need to be gradually increased, since the effects of interpolation naturally becomes stronger as learning rate decreases.

\begin{algorithm}[H]
\caption{Periodic Interpolation applied to Problem~\ref{eqn:toy_problem_expanded}}
\label{alg:pi}
\begin{algorithmic}[1]
    \Function{PI}{$\alpha$, $K$} 
        \For{$t = 1$ to $T$}
            \State $Q = s\cdot \lfloor 1 / s \cdot W^{(t-1)}\rceil$
            \State $G = X^T(XQ - XW^*)$ 
            \State perform optimizer step with gradient $G$ to obtain updated weight $W_{\text{curr}}^{(t)}$
            \If{$t$ \% $K = 0$}
                \State $W^{(t)} = (1 - \alpha) \cdot W_{\text{curr}}^{(t)} + \alpha \cdot (s\cdot \lfloor 1 / s \cdot W^{(t)}\rceil)$
            \Else
                \State $W^{(t)} = W_{\text{curr}}^{(t)}$
            \EndIf
        \EndFor
    \EndFunction
\end{algorithmic}
\end{algorithm}

\subsection{Other rounding Jacobian Estimators}\label{sec:toy_grad_est}
In Table~\ref{tab:toy_grad_est}, we additionally apply CEWT to multiple rounding Jacobian Estimators, demonstrating that CEWT can suppress weight oscillation and improve model performance regardless of the rounding Jacobian estimator used. 

\begin{table}[H]
\centering
\resizebox{0.8\textwidth}{!}{%
\begin{tabular}{|l|l|c|c|c|c|}
\hline
                     &           & STE & HTGE & JacQuant-Probe & EWGS \\ \hline
\multirow{2}{*}{MSE$(XQ, XW^*)$ (unit: 1e-3)} & Default & 81±0 & 53±0   & 63±0  &    61±0       \\  
                     & With CEWT & \textbf{45±0} & \textbf{46±0} & \textbf{52±0}  & \textbf{51±0}          \\ \hline
\multirow{2}{*}{RBM$(\text{vec}(1/s\cdot W))$ (unit: 1e-3)} & Default & 573±1 & 177±1 &  349±1 & 324±1      \\ 
                     & With CEWT & \textbf{10±0} & \textbf{10±0} & \textbf{10±0}  &  \textbf{10±0}         \\ \hline
\multirow{2}{*}{EMA Oscillation Freq (unit: 1e-4)} & Default & 701±2 & 245±1 &  511±2 & 392±1      \\ 
                     & With CEWT & \textbf{17±0} & \textbf{30±0} & \textbf{37±0}  &  \textbf{13±0}         \\ \hline
\end{tabular}%
}
\caption{Effectiveness of CEWT on different rounding Jacobian estimators.}
\label{tab:toy_grad_est}
\end{table}

\subsection{When is CEWT Less Effective?}\label{sec:sweep_dniter}
In this section, we discuss when rounding boundary weight oscillation is less severe, and when CEWT is less effective.

In Table~\ref{tab:toy_sweep_n}, we fix $d=2^{10}$ and the number of training iterations (200), and we sweep over $n \in \{1, 2, 4, 8, 16, 32, 64, 128, 256, 512, 1024\}$. Our results show that when the model size is significantly larger than the dataset size, rounding boundary weight oscillation is less severe, and the loss reduction effect of CEWT is small and almost negligible. 

In Table~\ref{tab:toy_sweep_iter}, we fix $n=2^{10}$ and $d=2^{10}$, and we sweep over the number of training iterations in $\{1, 3, 6, 12, 25, 50, 100, 200\}$. Our results show that while CEWT can increase MSE loss when the number of training iterations is small (e.g. less than 100), as long as the model is trained to full convergence (i.e. STE loss no longer decreases significantly when the number of epochs doubles), CEWT can consistently achieve lower loss than the STE. Table~\ref{tab:toy_sweep_iter} also suggests rounding boundary oscillation is less severe when the training time is not sufficiently long.

In Table~\ref{tab:toy_sweep_d}, we fix $n=2^{10}$ and the number of iterations (200), and sweep over $d \in \{1, 2, 4, 8, 16, 32, 64, 128, 256, 512, 1024\}$. Our result suggest CEWT can consistently reduce loss as long as model size is sufficiently large (e.g. $d \ge 32$). CEWT is designed for large models, as it relies on the law of large numbers to guarantee that that a random permutation of weights look indistinguishable from Gaussian noise. 

In conclusion, CEWT is most effective for large-scale Quantization-Aware Pre-Training (QAPT), or training models with \textbf{many low-precision parameters} for \textbf{a long duration} on a \textbf{large dataset}. 

\begin{table}[H]
\centering
\resizebox{\textwidth}{!}{%
\renewcommand{\arraystretch}{1.3} 
\begin{tabular}{|l|l|l|l|l|l|l|l|l|l|l|l|l|}
\hline
 & $n=$        & 1  & 2  & 4  & 8  & 16 & 32 & 64 & 128 & 256 & 512 & 1024 \\ \hline
MSE$(XQ, XW^*)$ (unit: 1e-3) & STE  & \textbf{1±0}  & \textbf{1±0}  & \textbf{1±0}  & \textbf{2±0}  & 3±0  & 4±0  & 7±0  & 13±0  & 23±0  & 44±0  & 81±0    \\
MSE$(XQ, XW^*)$ (unit: 1e-3) & CEWT & \textbf{1±0}  & \textbf{1±0}  & \textbf{1±0}  & \textbf{2±0}  & \textbf{2±0}  & \textbf{3±0}  & \textbf{5±0}  & \textbf{9±0}   & \textbf{14±0}  & \textbf{25±0}  & \textbf{45±0}   \\ \hline
RBM$(\text{vec}(1/s\cdot W))$ (unit: 1e-3) & STE  & 11±0 & 12±0 & 15±0 & 19±0 & 27±0 & 40±0 & 64±0 & 108±1 & 186±0 & 332±1 & 573±1  \\ 
RBM$(\text{vec}(1/s\cdot W))$ (unit: 1e-3) & CEWT & \textbf{10±0} & \textbf{10±0} & \textbf{10±0} & \textbf{10±0} & \textbf{10±0} & \textbf{10±0} & \textbf{10±0} & \textbf{10±0}  & \textbf{10±0}  & \textbf{10±0}  & \textbf{10±0}   \\ \hline
\end{tabular}%
}
\caption{Rounding boundary mass and MSE loss for different values of $n$ (dataset size).}
\label{tab:toy_sweep_n}
\end{table}

\begin{table}[H]
\centering
\resizebox{\textwidth}{!}{%
\renewcommand{\arraystretch}{1.3} 
\begin{tabular}{|l|l|l|l|l|l|l|l|l|l|l|}
\hline
 & Iterations =        & 1  & 3  & 6  & 12  & 25 & 50 & 100 & 200  \\ \hline
MSE$(XQ, XW^*)$ (unit: 1e-3) & STE  & 1847±10  & \textbf{1609±9}  & \textbf{1277±7}  & \textbf{747±5} & \textbf{244±1}  & \textbf{90±0}  & 82±0 & 81±0   \\
MSE$(XQ, XW^*)$ (unit: 1e-3) & CEWT & \textbf{1839±9}  & 1625±9  & 1343±7  & 859±6  & 281±1  & 107±0  & \textbf{63±0}  & \textbf{45±0}   \\ \hline
RBM$(\text{vec}(1/s\cdot W))$ (unit: 1e-3) & STE  & 11±0 & 12±0 & 15±0 & 19±0 & 27±0 & 40±0 & 64±0 & 108±1  \\ 
RBM$(\text{vec}(1/s\cdot W))$ (unit: 1e-3) & CEWT & \textbf{10±0} & \textbf{10±0} & \textbf{10±0} & \textbf{10±0} & \textbf{10±0} & \textbf{10±0} & \textbf{10±0} & \textbf{10±0} \\ \hline
\end{tabular}%
}
\caption{Rounding boundary mass and MSE loss for different number of training iterations.}
\label{tab:toy_sweep_iter}
\end{table}

\begin{table}[H]
\centering
\resizebox{\textwidth}{!}{%
\renewcommand{\arraystretch}{1.3} 
\begin{tabular}{|l|l|l|l|l|l|l|l|l|l|l|l|}
\hline
 & $d=$        & 2  & 4  & 8  & 16 & 32 & 64 & 128 & 256 & 512 & 1024 \\ \hline
MSE$(XQ, XW^*)$ (unit: 1e-3) & STE  & \textbf{398±452}  & \textbf{258±352}  & \textbf{130±47}  & \textbf{137±38}  & 128±37  & 128±6  & 122±2  & 117±1  & 105±1  & 81±0    \\
MSE$(XQ, XW^*)$ (unit: 1e-3) & CEWT & 1849±2115  & 1876±1619  & 313±281  & 117±41  & \textbf{95±8}  & \textbf{90±4}  & \textbf{85±1}   & \textbf{79±1}   & \textbf{66±0}   & \textbf{45±0}   \\ \hline
RBM$(\text{vec}(1/s\cdot W))$ (unit: 1e-3) & STE  & 600±671 & 775±244 & 756±114 & 753±52 & 741±40 & 733±21 & 712±5 & 693±2 & 661±1 & 573±1  \\ 
RBM$(\text{vec}(1/s\cdot W))$ (unit: 1e-3) & CEWT & \textbf{0±0} & \textbf{0±0} & \textbf{13±34} & \textbf{11±9} & \textbf{11±4} & \textbf{10±1} & \textbf{10±0}  & \textbf{10±0}  & \textbf{10±0}  & \textbf{10±0}   \\ \hline
\end{tabular}%
}
\caption{Rounding boundary mass and MSE loss for different values of $d$ (model size).}
\label{tab:toy_sweep_d}
\end{table}

\section{Ablation Studies on Removing the Hadamard Transform and TGCS Quantization}
In this section, we present our ablation studies in Table~\ref{tab:ablation} and the effect of TGCS on weight distribution in Figure~\ref{fig:ablation}.
\begin{table}[H]
\centering
\resizebox{\textwidth}{!}{%
\renewcommand{\arraystretch}{1.3} 
\begin{tabular}{|l|lll|l|l|l|l|}
\hline
\multirow{2}{*}{Row} & \multicolumn{3}{l|}{Weight Quantizer}                       & \multicolumn{1}{c|}{\multirow{2}{*}{Optimizer}} & \multicolumn{1}{c|}{\multirow{2}{*}{Perplexity}} & \multicolumn{1}{c|}{\multirow{2}{*}{RBM}} & \multicolumn{1}{c|}{\multirow{2}{*}{Notes}}          \\ \cline{2-4}
                     & \multicolumn{1}{l|}{Type}  & \multicolumn{1}{l|}{HT} & TGCS & \multicolumn{1}{c|}{}                           & \multicolumn{1}{c|}{}                            & 
                     \multicolumn{1}{c|}{} &\multicolumn{1}{c|}{}                                \\ \hline
1                    & \multicolumn{1}{l|}{QuEST} & \multicolumn{1}{l|}{T}  & F    & AdamH                                           & 32.42 & 0.079                                        & Vanilla QuEST without CEWT                           \\ \hline
2                    & \multicolumn{1}{l|}{QuEST} & \multicolumn{1}{l|}{F}  & F    & Adam-CEWT                                       & 31.53 &  0.071                                          & Introduce CEWT without TGCS and remove HT            \\ \hline
3                    & \multicolumn{1}{l|}{QuEST} & \multicolumn{1}{l|}{F}  & T    & Adam-CEWT                                       & \textbf{30.70} & \textbf{0.056}                                  & Further Introduce TGCS (Intended usage of this work) \\ \hline
4                    & \multicolumn{1}{l|}{QuEST} & \multicolumn{1}{l|}{T}  & T    & Adam-CEWT-HT                                    & 30.81 & \textbf{0.056}                                           & Introduce HT to both weight quantizer and optimizer  \\ \hline
5                    & \multicolumn{1}{l|}{BBQ}   & \multicolumn{1}{l|}{T}  & F    & AdamH                                           & 31.61 &  0.029                                          & Vanilla BBQ without CEWT                             \\ \hline
6                    & \multicolumn{1}{l|}{BBQ}   & \multicolumn{1}{l|}{F}  & F    & Adam-CEWT                                       & 30.69 & 0.015                                           & Introduce CEWT without TGCS and remove HT            \\ \hline
7                    & \multicolumn{1}{l|}{BBQ}   & \multicolumn{1}{l|}{F}  & T    & Adam-CEWT                                       & \textbf{30.13} & \textbf{0.010}                                  & Further Introduce TGCS (Intended usage of this work) \\ \hline
8                    & \multicolumn{1}{l|}{BBQ}   & \multicolumn{1}{l|}{T}  & T    & Adam-CEWT-HT                                    & 30.29 & \textbf{0.010}                                          & Introduce HT to both weight quantizer and optimizer  \\ \hline
\end{tabular}%
}
\caption{Evaluation Perplexity of HO, CEWT w.o TGCS, CEWT, and CEWT-HT.}
\label{tab:ablation}
\end{table}
\begin{figure}[H]
    \centering
    \begin{subfigure}[t]{0.49\textwidth}
        \includegraphics[width=\linewidth]{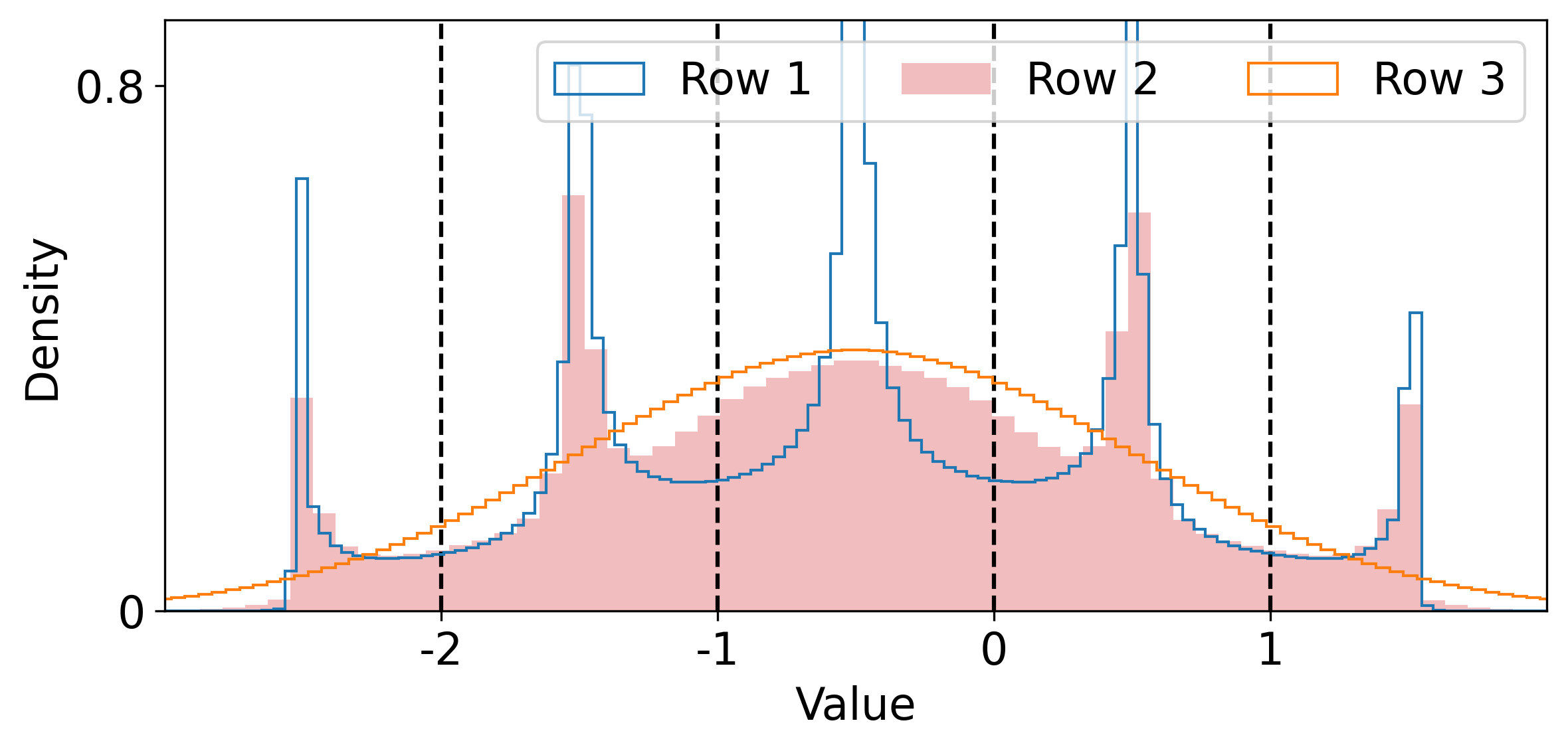}
        \caption{QuEST}
        \label{fig:ablation_a}
    \end{subfigure}
    \begin{subfigure}[t]{0.49\textwidth}
        \includegraphics[width=\linewidth]{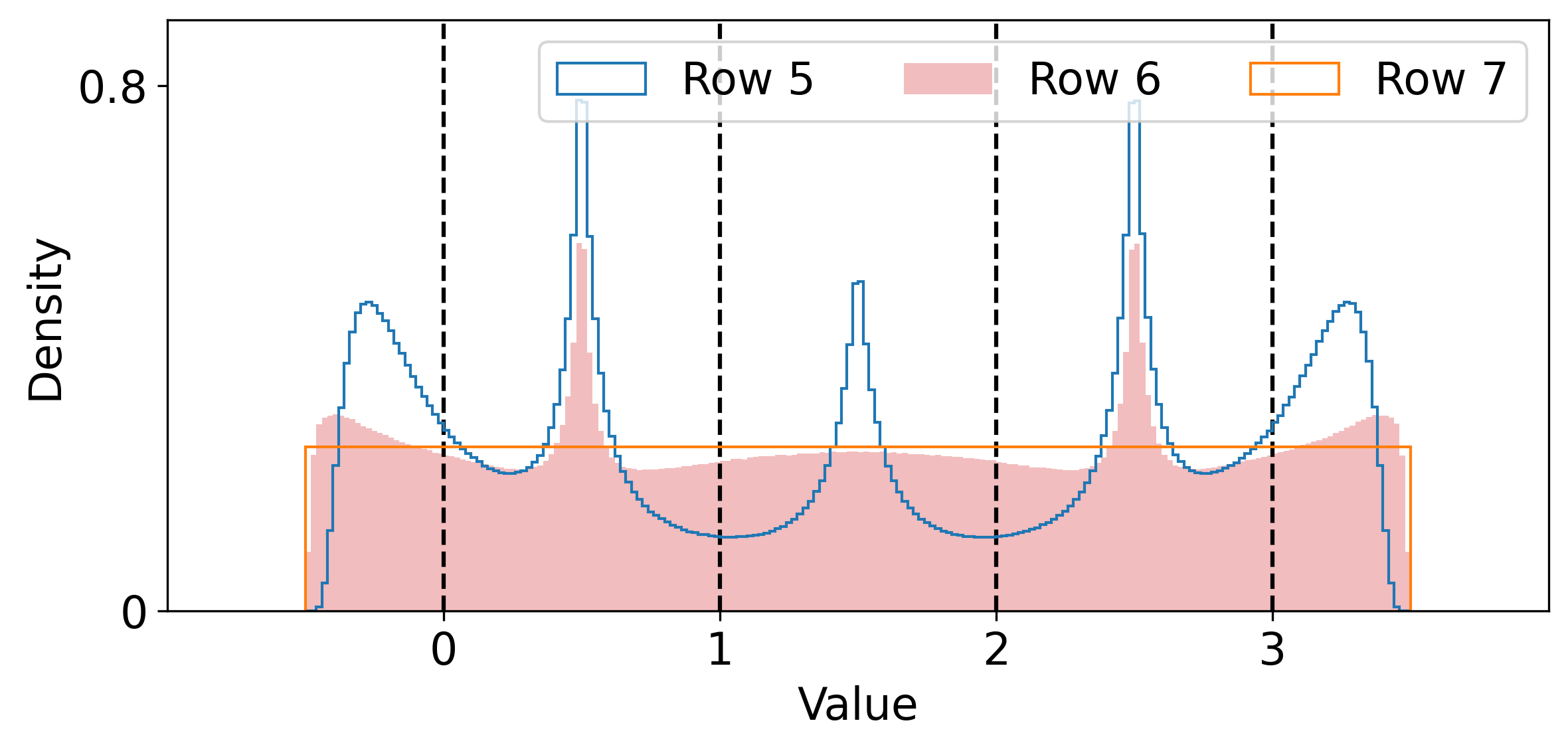}
        \caption{BBQ}
        \label{fig:ablation_b}
    \end{subfigure}
    \caption{Pre-clip (QuEST) and pre-round (BBQ) weight distribution of models (rows) in Table \ref{tab:ablation}.}
    \label{fig:ablation}
\end{figure}

\section{Example Hypersphere Optimizers}\label{sec:optimizer}
In this section, we show how a few example hypersphere optimizers can be integrated with CEWT. 
AdamH is a Frobenius-norm-hypersphere-optimization~\citep{hyperballOptimization2026} variant of Adam~\citep{kingma2017adammethodstochasticoptimization}, and its algorithm (with optional CEWT integration) is in Algorithm~\ref{alg:adamh}.
MuonH is a Frobenius-norm-hypersphere-optimization variant of Muon~\citep{muon}, and its algorithm (with optional CEWT integration) is in Algorithm~\ref{alg:muonh}.
LionH is a Frobenius-norm-hypersphere-optimization variant of Muon~\citep{muon}, and its algorithm (with optional CEWT integration) is in Algorithm~\ref{alg:muonh}. 
SSO~\citep{xie2026controlled} is a spectral-norm hypersphere optimizer, and its algorithm (with optional CEWT integration) is shown in Algorithm~\ref{alg:sso}. While conceptually SSO performs a \textbf{post-update} retraction back to the radius-$R$ hypersphere, its implementation (Algorithm~\ref{alg:sso}) uses a \textbf{pre-update} retraction step for efficiency purposes. To integrate CEWT with pre-update retraction, we add Line~\ref{line:ssoh_cewt_practical} to Algorithm~\ref{alg:sso}. Line~\ref{line:ssoh_cewt_practical} normalizes $\hat{W}$ such that the Gaussianization does not drastically change the magnitude of weights. 

\begin{algorithm}[H]
\caption{AdamH with optional CEWT Integration for Matrix Weights $W^{(t-1)} \in \mathbb{R}^{m\times n}$ \\ \colorbox{yellow!30}{Adam (Lines \ref{line:adam_begin} to \ref{line:adam_end})} \colorbox{red!20}{HO (Lines \ref{line:adamh_ho_r}, \ref{line:adamh_ho_u}, \ref{line:adamh_ho_a}, \ref{line:adamh_ho_b}, and \ref{line:adamh_ho_c})} \colorbox{blue!20}{CEWT (Line \ref{line:adamh_cewt})} }
\label{alg:adamh}
\begin{algorithmic}[1]
    \Function{AdamH}{$\beta_1=0.9$, $\beta_2=0.999$, $\epsilon=10^{-8}$, $\text{integrate\_cewt}=\text{False}$}
        \State \colorbox{red!20}{$R = ||W^{(0)}||_F$}\label{line:adamh_ho_r}
        \State \colorbox{yellow!30}{$M^{(0)} = 0$}\label{line:adam_begin}
        \State \colorbox{yellow!30}{$V^{(0)} = 0$}
        \For{$t = 1$ to $T$}
            \State \colorbox{yellow!30}{Denote $G^{(t)}$ as a stochastic gradient of loss with respect to $W^{(t-1)}$} 
            \State \colorbox{yellow!30}{$M^{(t)} = \beta_1 \cdot M^{(t-1)} + (1 - \beta_1)\cdot G^{(t)}$}
            \State \colorbox{yellow!30}{$V^{(t)} = \beta_1 \cdot V^{(t-1)} + (1 - \beta_1)\cdot (G^{(t)} \odot G^{(t)})$} \Comment{Element-wise multiplication ($\odot$)}
            \State \colorbox{yellow!30}{$\hat{M}^{(t)} = 1 / (1 - (\beta_1)^t)\cdot M^{(t)}$}
            \State \colorbox{yellow!30}{$\hat{V}^{(t)} = 1 / (1 - (\beta_2)^t)\cdot V^{(t)}$}
            \State \colorbox{yellow!30}{$O^{(t)} = \hat{M}^{(t)} \oslash (\sqrt{\hat{V}^{(t)}} + \epsilon)$} \Comment{Element-wise square root ($\sqrt{\cdot}$) and division ($\oslash$)} \label{line:adam_end}
            \State \colorbox{red!20}{$U^{(t)} = 1 / ||O^{(t)}||_F \cdot O^{(t)}$} \Comment{Ensures $||U^{(t)}||_F = 1$} \label{line:adamh_ho_u} 
            \State \colorbox{red!20}{$\bar{W}^{(t)} = W^{(t-1)} - \eta^{(t)} R U^{(t)}$} \Comment{Equation~\ref{eqn:hyperball_a}} \label{line:adamh_ho_a}
            \If{integrate\_cewt}
                \State \colorbox{blue!20}{$\hat{W}^{(t)} = \text{mat}\left(\Phi^{-1}\left(\frac{1}{mn}\cdot\left(\text{ranks}(\text{vec}(\bar{W}^{(t)})) + \frac{1}{2}\right)\right)\right)$}\Comment{Equation~\ref{eqn:cewt}} \label{line:adamh_cewt}
            \Else
                \State \colorbox{red!20}{$\hat{W}^{(t)} = \bar{W}^{(t)}$}\Comment{Equation~\ref{eqn:hyperball_b}} \label{line:adamh_ho_b}
            \EndIf
            \State \colorbox{red!20}{$W^{(t)} = \frac{R}{||\hat{W}^{(t)}||_F}\hat{W}^{(t)}$}\Comment{Equation~\ref{eqn:hyperball_c}} \label{line:adamh_ho_c}
        \EndFor
    \EndFunction
\end{algorithmic}
\end{algorithm}

\begin{algorithm}[H]
\caption{MuonH with optional CEWT Integration for Matrix Weights $W^{(t-1)} \in \mathbb{R}^{m\times n}$ \\ \colorbox{yellow!30}{Muon (Lines \ref{line:muon_begin} to \ref{line:muon_end})} \colorbox{red!20}{HO (Lines \ref{line:muonh_ho_r}, \ref{line:muonh_ho_u}, \ref{line:muonh_ho_a}, \ref{line:muonh_ho_b}, and \ref{line:muonh_ho_c})} \colorbox{blue!20}{CEWT (Line \ref{line:muonh_cewt})} }
\label{alg:muonh}
\begin{algorithmic}[1]
    \Function{MuonH}{$\mu=0.95$, $\text{nesterov}=\text{True}$, $\text{integrate\_cewt}=\text{False}$}
        \State \colorbox{red!20}{$R = ||W^{(0)}||_F$} \label{line:muonh_ho_r}
        \State \colorbox{yellow!30}{$B^{(0)} = 0$}\label{line:muon_begin}
        \For{$t = 1$ to $T$}
            \State \colorbox{yellow!30}{Denote $G^{(t)}$ as a stochastic gradient of loss with respect to $W^{(t-1)}$} 
            \State \colorbox{yellow!30}{$B^{(t)} = \mu \cdot B^{(t-1)} + G^{(t)}$}
            \If{nesterov}
                \State \colorbox{yellow!30}{$O^{(t)} = \text{NewtonShultz}(\mu \cdot B^{(t)} + G^{(t)})$}
            \Else
                \State \colorbox{yellow!30}{$O^{(t)} = \text{NewtonShultz}(B^{(t)})$}  
            \EndIf \label{line:muon_end}
            \State \colorbox{red!20}{$U^{(t)} = 1 / ||O^{(t)}||_F \cdot O^{(t)}$}  \Comment{Ensures $||U^{(t)}||_F = 1$} \label{line:muonh_ho_u}
            \State \colorbox{red!20}{$\bar{W}^{(t)} = W^{(t-1)} - \eta^{(t)} R U^{(t)}$} \Comment{Equation~\ref{eqn:hyperball_a}} \label{line:muonh_ho_a}
            \If{integrate\_cewt}
                \State \colorbox{blue!20}{$\hat{W}^{(t)} = \text{mat}\left(\Phi^{-1}\left(\frac{1}{mn}\cdot\left(\text{ranks}(\text{vec}(\bar{W}^{(t)})) + \frac{1}{2}\right)\right)\right)$}\Comment{Equation~\ref{eqn:cewt}} \label{line:muonh_cewt}
            \Else
                \State \colorbox{red!20}{$\hat{W}^{(t)} = \bar{W}^{(t)}$}\Comment{Equation~\ref{eqn:hyperball_b}} \label{line:muonh_ho_b}
            \EndIf
            \State \colorbox{red!20}{$W^{(t)} = \frac{R}{||\hat{W}^{(t)}||_F}\hat{W}^{(t)}$}\Comment{Equation~\ref{eqn:hyperball_c}} \label{line:muonh_ho_c}
        \EndFor
    \EndFunction
\end{algorithmic}
\end{algorithm}

\begin{algorithm}[H]
\caption{LionH with optional CEWT Integration for Matrix Weights $W^{(t-1)} \in \mathbb{R}^{m\times n}$ \\ \colorbox{yellow!30}{Lion (Lines \ref{line:lion_begin} to \ref{line:lion_end})} \colorbox{red!20}{HO (Lines \ref{line:lionh_ho_r}, \ref{line:lionh_ho_u}, \ref{line:lionh_ho_a}, \ref{line:lionh_ho_b}, and \ref{line:lionh_ho_c})} \colorbox{blue!20}{CEWT (Line \ref{line:lionh_cewt})} }
\label{alg:lionh}
\begin{algorithmic}[1]
    \Function{LionH}{$\beta_1=0.9$, $\beta_2= 0.99$, $\text{integrate\_cewt}=\text{False}$}
        \State \colorbox{red!20}{$R = ||W^{(0)}||_F$} \label{line:lionh_ho_r}
        \State \colorbox{yellow!30}{$M^{(0)} = 0$}\label{line:lion_begin}
        \For{$t = 1$ to $T$}
            \State \colorbox{yellow!30}{Denote $G^{(t)}$ as a stochastic gradient of loss with respect to $W^{(t-1)}$} 
            \State \colorbox{yellow!30}{$M^{(t)} = \beta_2 \cdot M^{(t-1)} + (1 - \beta_2) \cdot G^{(t)}$}
            \State \colorbox{yellow!30}{$O^{(t)} = \text{sign}(\beta_1 \cdot M^{(t-1)} + (1 - \beta_1) \cdot G^{(t)})$}\Comment{Element-wise sign function}\label{line:lion_end}
            \State \colorbox{red!20}{$U^{(t)} = 1 / ||O^{(t)}||_F \cdot O^{(t)}$}  \Comment{Ensures $||U^{(t)}||_F = 1$} \label{line:lionh_ho_u}
            \State \colorbox{red!20}{$\bar{W}^{(t)} = W^{(t-1)} - \eta^{(t)} R U^{(t)}$} \Comment{Equation~\ref{eqn:hyperball_a}} \label{line:lionh_ho_a}
            \If{integrate\_cewt}
                \State \colorbox{blue!20}{$\hat{W}^{(t)} = \text{mat}\left(\Phi^{-1}\left(\frac{1}{mn}\cdot\left(\text{ranks}(\text{vec}(\bar{W}^{(t)})) + \frac{1}{2}\right)\right)\right)$}\Comment{Equation~\ref{eqn:cewt}} \label{line:lionh_cewt}
            \Else
                \State \colorbox{red!20}{$\hat{W}^{(t)} = \bar{W}^{(t)}$}\Comment{Equation~\ref{eqn:hyperball_b}} \label{line:lionh_ho_b}
            \EndIf
            \State \colorbox{red!20}{$W^{(t)} = \frac{R}{||\hat{W}^{(t)}||_F}\hat{W}^{(t)}$}\Comment{Equation~\ref{eqn:hyperball_c}} \label{line:lionh_ho_c}
        \EndFor
    \EndFunction
\end{algorithmic}
\end{algorithm}

\begin{algorithm}[H]
\caption{SSO with optional CEWT Integration for Matrix Weights $\mathbf{\hat{W}^{(t-1)}} \in \mathbb{R}^{m\times n}$ \\ \colorbox{yellow!30}{Update Computation (Lines \ref{line:sso_begin} to \ref{line:sso_end})} \colorbox{red!20}{HO (Lines \ref{line:sso_ho_r}, \ref{line:ssoh_ho_a}, \ref{line:ssoh_ho_b}, and \ref{line:ssoh_ho_c})} \colorbox{blue!20}{CEWT (Line \ref{line:ssoh_cewt} and \ref{line:ssoh_cewt_practical})} }
\label{alg:sso}
\begin{algorithmic}[1]
    \Function{SSO}{$\beta=0.9$, $\epsilon=10^{-6}$, $\text{integrate\_cewt}=\text{False}$}
        \State \colorbox{red!20}{$R = \sqrt{m / n}$}\label{line:sso_ho_r}
        \State \colorbox{yellow!30}{$M^{(0)} = 0$}\label{line:sso_begin}
        \For{$t = 1$ to $T$}
            \State \colorbox{yellow!30}{Denote $G^{(t)}$ as a stochastic gradient of loss with respect to $\mathbf{\hat{W}^{(t-1)}}$} 
            \State \colorbox{yellow!30}{$M^{(t)} = \beta \cdot M^{(t-1)} + (1 - \beta) \cdot G^{(t)}$}
            \State \colorbox{yellow!30}{$\hat{M}^{(t)} = 1 / ||M^{(t)}||_F \cdot M^{(t)}$}
            \State \colorbox{yellow!30}{$\sigma^{(t-1)}, u^{(t)}, v^{(t)} = \text{PowerIteration}(\hat{W}^{(t-1)})$}\Comment{$\sigma^{(t-1)}\approx||\hat{W}^{(t-1)}||_2$}
            \State \colorbox{yellow!30}{$\Theta^{(t)} = u^{(t)}v^{(t)}$}
            \State \colorbox{red!20}{$W^{(t-1)} = \frac{R}{\sigma^{(t-1)}}\hat{W}^{(t-1)}$}\Comment{Equation~\ref{eqn:hyperball_c} implemented as pre-update retraction} \label{line:ssoh_ho_c}
            \State \colorbox{yellow!30}{Define $h(\lambda)=\langle \Theta^{(t)}, \text{msign}(\hat{M}^{(t)}  + \lambda \Theta^{(t)}) \rangle$}
            \State \colorbox{yellow!30}{$(\lambda^*)^{(t)} = \text{Bisection}(h, \text{tolerance}=\epsilon)$}
            \State \colorbox{yellow!30}{$U^{(t)} = \text{msign}\left(\hat{M}^{(t)}  + (\lambda^*)^{(t)} \Theta^{(t)}\right)$} \Comment{Note: $||U^{(t)}||_2 = 1$}\label{line:sso_end}
            \State \colorbox{red!20}{$\bar{W}^{(t)} = W^{(t-1)} - \eta^{(t)} R U^{(t)}$} \Comment{Equation~\ref{eqn:hyperball_a}} \label{line:ssoh_ho_a}
            \If{integrate\_cewt}
                \State \colorbox{blue!20}{$\hat{W}_{\text{temp}}^{(t)} = \text{mat}\left(\Phi^{-1}\left(\frac{1}{mn}\cdot\left(\text{ranks}(\text{vec}(\bar{W}^{(t)})) + \frac{1}{2}\right)\right)\right)$}\Comment{Equation~\ref{eqn:cewt}} \label{line:ssoh_cewt}
                \State \colorbox{blue!20}{$\hat{W}^{(t)} = \frac{||\bar{W}^{(t)}||_F}{||\hat{W}_{\text{temp}}^{(t)}||_F} \cdot \hat{W}_{\text{temp}}^{(t)}$}\Comment{Ensure $||\hat{W}^{(t)}||_F=||\bar{W}^{(t)}||_F$ } \label{line:ssoh_cewt_practical}
            \Else
                \State \colorbox{red!20}{$\hat{W}^{(t)} = \bar{W}^{(t)}$}\Comment{Equation~\ref{eqn:hyperball_b}} \label{line:ssoh_ho_b}
            \EndIf
        \EndFor
    \EndFunction
\end{algorithmic}
\end{algorithm}

\pagebreak
\section{Example Quantizers}\label{sec:quantizer}
In this section we present the algorithm of a few quantizers and how they can be integrated with CEWT (Hadamard Transform Removal and TGCS). 
BBQ~\citep{yang2026boosting} is a per-channel quantizer with Hadamard Transform, and its algorithm (with optional CEWT integration) is shown in Algorithm~\ref{alg:bbq}.
QuEST~\citep{panferov2025queststabletrainingllms} is a per-channel quantizer with Hadamard Transform, and its algorithm (with optional CEWT integration) is shown in Algorithm~\ref{alg:quest}.
LSQ~\citep{esser2020learnedstepsizequantization} is a per-tensor quantizer, and its algorithm is shown in Algorithm~\ref{alg:lsq}. LSQ does not require customization when integrated with CEWT.

{
\algrenewcommand\textproc[1]{#1}
\begin{algorithm}[H]
\caption{BBQ with optional CEWT Integration for Matrix Weights $W^{(t)} \in \mathbb{R}^{m\times n}$ \\ \colorbox{orange!20}{HT Removal (Line \ref{line:bbq_ht_removal})} \colorbox{blue!20}{TGCS (Line \ref{line:bbq_tgcs})} \colorbox{green!40!black!20}{Pre-clip and Pre-round Weights (Lines \ref{line:bbq_preclip} and \ref{line:bbq_preround})}}
\label{alg:bbq}
\begin{algorithmic}[1]
    \State 
    \Function{quant\_scale}{$v \in \mathbb{R}^d$}
        \State return $\sqrt{\frac{1}{d}\sum_{i=0}^{d-1}(v_i)^2}$
    \EndFunction
    \State
    \Function{dequant\_scale}{$v \in \mathbb{R}^d$}
        \State return $\zeta^*/(2^b)\cdot\sqrt{\frac{1}{d}\sum_{i=0}^{d-1}(v_i)^2}$\Comment{$\zeta^*$ is a constant defined in the original paper.}
    \EndFunction
    \State
    \Function{clipround}{$M \in \mathbb{R}^{m \times n}$}
        \State \colorbox{green!40!black!20}{preclip = $M$} \label{line:bbq_preclip}
        \State \colorbox{green!40!black!20}{preround = $2^b \cdot \Phi(\text{preclip}) - \frac{1}{2}$} \Comment{Element-wise standard Gaussian CDF ($\Phi$)} \label{line:bbq_preround}
        \State return $\lfloor \text{preround} \rceil - 2^{b-1} + \frac{1}{2}$ \Comment{Element-wise round-to-nearest-integer ($\lfloor \cdot \rceil$)}
    \EndFunction
    \State
    \Function{HT}{$M \in \mathbb{R}^{m \times n}$}
        \State Denote $2^k$ as the Hadamard block size
        \State Denote $H_k$ as the Hadamard matrix with shape $2^k \times 2^k$
        \State assert n \% $2^k$ = 0
        \State $\bar{M}$ = M.reshape($m\cdot n/2^k$, $2^k$)
        \State $\hat{M} = \bar{M} H_k$
        \State return $\hat{M}$.reshape(m, n)
    \EndFunction
    \State
    \Function{BBQ}{$W \in \mathbb{R}^{m \times n}$, $\text{integrate\_cewt} = \text{False}$}
        \If{\colorbox{orange!20}{not integrate\_cewt}} \Comment{See Section~\ref{sec:cewt_ht}}\label{line:bbq_ht_removal}
            \State W = HT(W)
        \EndIf
        \State Denote $Q \in \mathbb{R}^{m \times n}$ as quantized weights
        \For{$i = 0$ to $m - 1$}
            \If{integrate\_cewt}
                \State \colorbox{blue!20}{$s_q$ = quant\_scale$(\text{vec}(W))$} \Comment{See Section~\ref{sec:cewt_tgcs}}\label{line:bbq_tgcs}
            \Else
                \State $s_q$ = quant\_scale$(W_i)$
            \EndIf
            \State $s_{di} = \text{dequant\_scale}(W_i)$
            \State $Q_i = s_{di} \cdot \text{clipround}(1 / s_q \cdot W_i)$
        \EndFor
        \State return $Q$
    \EndFunction 
\end{algorithmic}
\end{algorithm}
}

{
\algrenewcommand\textproc[1]{#1}
\begin{algorithm}[H]
\caption{QuEST with optional CEWT Integration for Matrix Weights $W^{(t)} \in \mathbb{R}^{m\times n}$\\\colorbox{orange!20}{HT Removal (Line \ref{line:quest_ht_removal})} \colorbox{blue!20}{TGCS (Line \ref{line:quest_tgcs})} \colorbox{green!40!black!20}{Pre-clip and Pre-round Weights (Lines \ref{line:quest_preclip} and \ref{line:quest_preround})} }
\label{alg:quest}
\begin{algorithmic}[1]
    \State
    \Function{quant\_scale}{$v \in \mathbb{R}^d$}
        \State return $2\alpha^*/(2^b-1) \cdot \sqrt{\frac{1}{d}\sum_{i=0}^{d-1}(v_i)^2}$ \Comment{$\alpha^*$ is a constant defined in the original paper.}
    \EndFunction
    \State
    \Function{dequant\_scale}{$v \in \mathbb{R}^d$}
        \State return $2\alpha^*/(2^b-1) \cdot \sqrt{\frac{1}{d}\sum_{i=0}^{d-1}(v_i)^2}$
    \EndFunction
    \State
    \Function{clipround}{$M \in \mathbb{R}^{m \times n}$}
        \State \colorbox{green!40!black!20}{preclip = $M - \frac{1}{2}$} \label{line:quest_preclip}
        \State \colorbox{green!40!black!20}{preround = $\text{clip}(\text{preclip}, -2^{b-1}, 2^{b-1} - 1)$} \Comment{Element-wise clipping} \label{line:quest_preround}
        \State return $\lfloor \text{preround} \rceil + \frac{1}{2}$ \Comment{Element-wise round-to-nearest-integer ($\lfloor \cdot \rceil$)}
    \EndFunction
    \State
    \Function{HT}{$M \in \mathbb{R}^{m \times n}$}
        \State Denote $2^k$ as the Hadamard block size
        \State Denote $H_k$ as the Hadamard matrix with shape $2^k \times 2^k$
        \State assert n \% $2^k$ = 0
        \State $\bar{M}$ = M.reshape($m\cdot n/2^k$, $2^k$)
        \State $\hat{M} = \bar{M} H_k$
        \State return $\hat{M}$.reshape(m, n)
    \EndFunction
    \State
    \Function{QuEST}{$W \in \mathbb{R}^{m \times n}$, $\text{integrate\_cewt} = \text{False}$}
        \If{\colorbox{orange!20}{not integrate\_cewt}}\Comment{See Section~\ref{sec:cewt_ht}}\label{line:quest_ht_removal}
            \State W = HT(W)
        \EndIf
        \State Denote $Q \in \mathbb{R}^{m \times n}$ as quantized weights
        \For{$i = 0$ to $m - 1$}
            \If{integrate\_cewt}
                \State \colorbox{blue!20}{$s_q$ = quant\_scale$(\text{vec}(W))$} \Comment{See Section~\ref{sec:cewt_tgcs}}\label{line:quest_tgcs}
            \Else
                \State $s_q$ = quant\_scale$(W_i)$
            \EndIf
            \State $s_{di} = \text{dequant\_scale}(W_i)$
            \State $Q_i = s_{di} \cdot \text{clipround}(1 / s_q \cdot W_i)$
        \EndFor
        \State return $Q$
    \EndFunction
\end{algorithmic}
\end{algorithm}
}

{
\algrenewcommand\textproc[1]{#1}
\begin{algorithm}[H]
\caption{LSQ for Matrix Weights $W^{(t)} \in \mathbb{R}^{m\times n}$ \\
\colorbox{green!40!black!20}{Pre-clip and Pre-round Weights (Lines \ref{line:lsq_preclip} and \ref{line:lsq_preround})}}
\label{alg:lsq}
\begin{algorithmic}[1]
    \State
    \Function{quant\_scale}{$v \in \mathbb{R}^d$}
        \State return $s$\Comment{$s$ is a learnable parameter independent of $v$}
    \EndFunction
    \State
    \Function{dequant\_scale}{$v \in \mathbb{R}^d$}
        \State return $s$\Comment{$s$ is a learnable parameter independent of $v$}
    \EndFunction
    \State
    \Function{clipround}{$M \in \mathbb{R}^{m \times n}$}
        \State \colorbox{green!40!black!20}{preclip = $M - \frac{1}{2}$} \label{line:lsq_preclip}
        \State \colorbox{green!40!black!20}{preround = $\text{clip}(\text{preclip}, -2^{b-1}, 2^{b-1} - 1)$} \label{line:lsq_preround} \Comment{Element-wise clipping} 
        \State return $\lfloor \text{preround} \rceil + \frac{1}{2}$ \Comment{Element-wise round-to-nearest-integer ($\lfloor \cdot \rceil$)}
    \EndFunction
    \State
    \Function{LSQ}{$W \in \mathbb{R}^{m \times n}$, $\text{integrate\_cewt} = \text{False}$}
        \State Denote $Q \in \mathbb{R}^{m \times n}$ as quantized weights
        \For{$i = 0$ to $m - 1$}
            \State $s_q$ = quant\_scale$(\text{vec}(W))$
            \State $s_{d} = \text{dequant\_scale}(\text{vec}(W))$
            \State $Q_i = s_{d} \cdot \text{clipround}(1 / s_q \cdot W_i)$
        \EndFor
        \State return $Q$
    \EndFunction
\end{algorithmic}
\end{algorithm}
}

\pagebreak
\section{Periodic Interpolation vs. Dampening Loss vs. CEWT (Deep Networks)}\label{sec:pi_dl_cewt}
Section~\ref{sec:cewt_alg} noted CEWT can be interpreted as a hyperparameter-free variant of Periodic Interpolation or Dampening loss. In Figure~\ref{fig:sweep_pi}, we compare the STE, CEWT, and Periodic Interpolation by setting the period to 1 iteration and sweeping through the interpolation ratio $\alpha$. Results suggest CEWT can achieve lower perplexity than Periodic Interpolation on 7 of the 8 chosen $\alpha$'s, while matching the performance of Periodic Interpolation with the best $\alpha$. In Figure~\ref{fig:sweep_dl}, we compare the STE, CEWT, and Dampening Loss by sweeping through the dampening hyperparameter $\alpha$. Results suggest CEWT can achieve lower perplexity than Dampening Loss on all 8 of the chosen hyperparameters.
\begin{figure}[H]
    \centering
    \includegraphics[width=\textwidth]{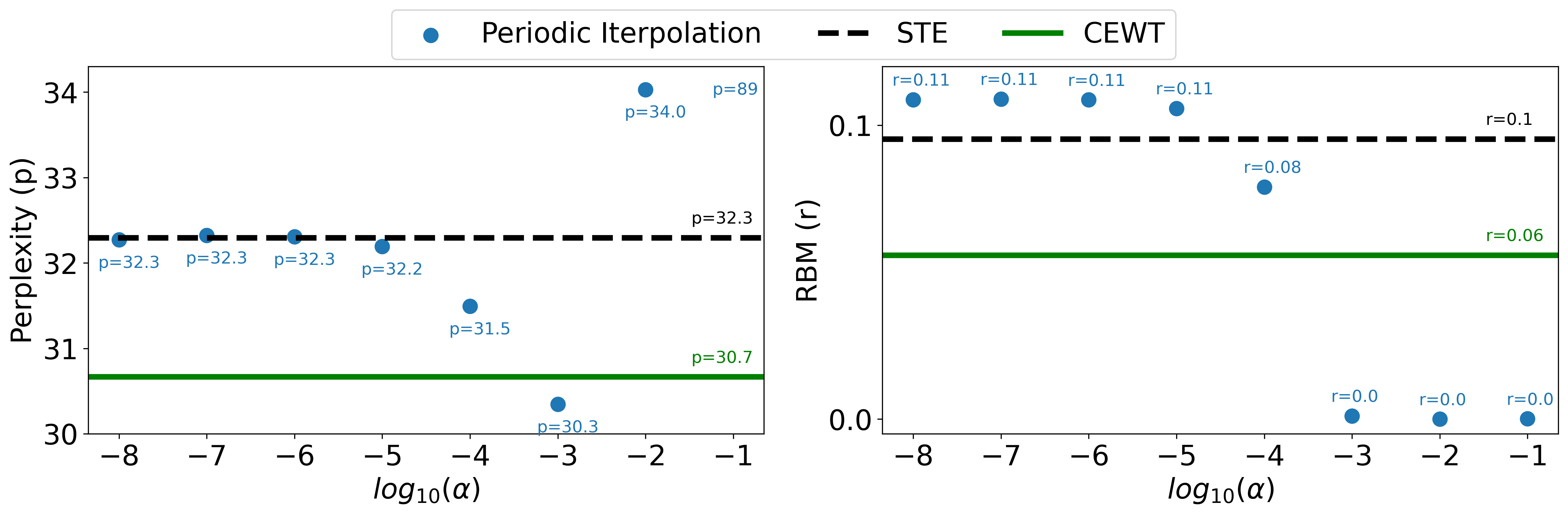}
    \caption{Comparing STE, CEWT, and Periodic Interpolation with a period of 1 and with grid search of the interpolation ratio hyperparameter $\alpha$. Experiments are conducted on a LLaMA-95M with activations and weights quantized to 2-bit (with QuEST) and pre-trained on 3 billion C4 tokens with the AdamH/Adam-CEWT optimizer. }
    \label{fig:sweep_pi}
\end{figure}
\begin{figure}[H]
    \centering
    \includegraphics[width=\textwidth]{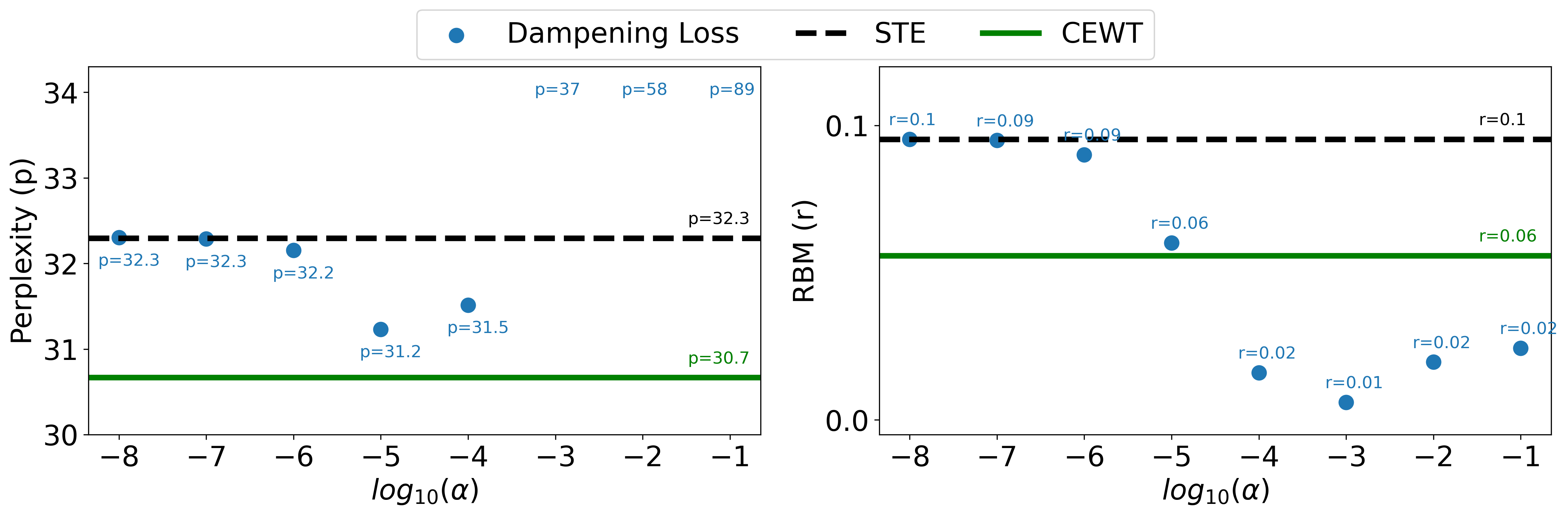}
    \caption{Comparing STE, CEWT, and Dampening Loss with a grid search of the dampening hyperparameter $\alpha$. Experiments are conducted on a LLaMA-95M with activations and weights quantized to 2-bit (with QuEST) and pre-trained on 3 billion C4 tokens with the AdamH/Adam-CEWT optimizer. }
    \label{fig:sweep_dl}
\end{figure}

\pagebreak
\section{Pseudo-Quantization Noise vs. CEWT (Deep Networks)}\label{sec:pqn}
In this section, we compare STE, PQN, and CEWT with a LLaMA-95M pre-trained on 3 billion C4 tokens. Table~\ref{tab:pqn} presents our results. We find that while both PQN and CEWT are capable of suppressing rounding boundary weight oscillation, PQN cannot consistently outperform STE while CEWT can consistently outperform both the STE and PQN.
\begin{table}[H]
\centering
\resizebox{0.7\textwidth}{!}{%
\renewcommand{\arraystretch}{1.5} 
\begin{tabular}{|l|l|cc|cc|cc|}
\hline
\multirow{2}{*}{Quantizer} & \multirow{2}{*}{Bits W/A} & \multicolumn{2}{l|}{AdamH + STE}     & \multicolumn{2}{l|}{AdamH + PQN} & \multicolumn{2}{l|}{Adam-CEWT} \\ 
                           &                           & \multicolumn{1}{l|}{PPL} & RBM & \multicolumn{1}{l|}{PPL}  & RBM  & \multicolumn{1}{l|}{PPL} & RBM \\ \hline

QuEST & 1 & 52.85 & 0.240 & 50.61 & \textbf{0.002} & \textbf{48.02} & 0.119 \\ \hline
BBQ & 1 & 51.06 & 0.034 & 49.54 & \textbf{0.001} & \textbf{46.12} & 0.010 \\ \hline
QuEST & 2 & 32.30 & 0.095 & 32.14 & \textbf{0.007} & \textbf{30.67} & 0.056 \\ \hline
BBQ & 2 & 31.80 & 0.023 & 31.75 & \textbf{0.005} & \textbf{30.18} & 0.010 \\ \hline
QuEST & 3 & 27.27 & 0.043 & 27.31 & \textbf{0.010} & \textbf{26.49} & 0.029 \\ \hline
BBQ & 3 & 27.35 & 0.017 & 27.64 & \textbf{0.008} & \textbf{26.67} & 0.010 \\ \hline
\end{tabular}%
}
\caption{Comparing STE, PQN, and CEWT.}
\label{tab:pqn}
\end{table}

\section{Full-Precision Perplexity and Hypersphere Learning Rates}\label{sec:full}
We tune our hypersphere optimizer learning rates by training a full-precision LLaMA-95M with the recipe used by 
\citet{panferov2025queststabletrainingllms}, log the ratio $||\Delta W||/||W||$, and set the hypersphere learning rate to the largest observed ratio. For larger models, we half our hypersphere learning rate whenever \citet{panferov2025queststabletrainingllms} halves the learning rate. 
Table~\ref{tab:fp_perp} presents the full-precision (without any quantization) perplexity of LLaMA models in Table~\ref{tab:main} trained on AdamW and AdamH, along with the learning rate used for both full-precision and quantized models. For GPT model experiments, we reuse the learning rates in Table~\ref{tab:fp_perp}, which are tuned for LLaMA models.
\begin{table}[H]
\centering
\resizebox{0.7\textwidth}{!}{%
\renewcommand{\arraystretch}{1.5} 
\begin{tabular}{|l|l|l|l|l|l|}
\hline
Model & Size & Optimizer & LR & Bits W/A & Perplexity \\ \hline
LLaMA & 95M  & AdamW (wd=0.1) & 0.0012   & 32/16 & 24.76         \\ \hline
LLaMA & 95M  & AdamH     &  0.01  & 32/16 & 23.95           \\ \hline
LLaMA & 127M  & AdamW (wd=0.1) & 0.0012   & 32/16 & 21.53         \\ \hline
LLaMA & 127M  & AdamH     &  0.01  & 32/16 & 20.86           \\ \hline
LLaMA & 206M  & AdamW (wd=0.1) & 0.0006   & 32/16 & 17.93         \\ \hline
LLaMA & 206M  & AdamH     &  0.005  & 32/16 & 17.52           \\ \hline
LLaMA & 332M  & AdamW (wd=0.1) & 0.0003   & 32/16 & -         \\ \hline
LLaMA & 332M  & AdamH     &  0.0025  & 32/16 & 15.17           \\ \hline
LLaMA & 610M  & AdamW (wd=0.1) & 0.00015   & 32/16 & -         \\ \hline
LLaMA & 610M  & AdamH     &  0.00125  & 32/16 & -           \\ \hline
\end{tabular}%
}
\caption{Full-Precision Perplexity and learning rate of LLaMA models in Table~\ref{tab:main}. A ``-'' indicates an experiment omitted due to resource constraint. }
\label{tab:fp_perp}
\end{table}

\pagebreak
\section{Zero-shot Evaluation}\label{sec:zero_shot}
Table~\ref{tab:zero_shot} presents our zero-shot results corresponding to the models in Table~\ref{tab:main} with more than 200M parameters. 
\begin{table}[H]
\centering
\resizebox{\textwidth}{!}{%
\renewcommand{\arraystretch}{1.5} 
\begin{tabular}{|l|l|l|c|c|c|c|c|c|c|}
\hline
M-S & Quant-Bits & Optimizer & HSWAG ($\uparrow$) & ARC-e ($\uparrow$) & ARC-c ($\uparrow$) & PiQA ($\uparrow$) & Winog ($\uparrow$) & Wiki ($\downarrow$) & Pile 10k ($\downarrow$) \\ \hline
L-206 & None-16 & AdamW & 30.12  & 41.29  & 17.92  & 65.29  & 50.67  & 48.90  & 539.96 \\ \hline
L-206 & None-16 & AdamH & 30.55  & 43.98  & 18.17  & 64.69  & 49.80  & 47.20  & 538.48 \\ \hline
L-332 & None-16 & AdamH & 32.69  & 46.09  & 19.45  & 67.14  & 51.38  & 37.92  & 407.21 \\ \hline
L-206 & BBQ-1 & MuonH     & 27.13  & \textbf{32.70}  & 17.66  & 56.64  & 49.57  & 128.73  & \textbf{2277.11} \\
L-206 & BBQ-1 & Muon-CEWT & \textbf{27.43}  & 32.24  & \textbf{18.77}  & \textbf{57.89}  & \textbf{53.35}  & \textbf{115.87}  & 2363.58 \\ \hline
L-206 & BBQ-1 & AdamH     & 27.12  & \textbf{33.16}  & 17.41  & 56.80  & \textbf{50.67}  & 133.04  & 2362.53 \\
L-206 & BBQ-1 & Adam-CEWT & \textbf{27.18}  & 33.12  & \textbf{19.45}  & \textbf{58.32}  & 50.51  & \textbf{115.07}  & \textbf{2043.74} \\ \hline
L-206 & BBQ-1 & LionH     & 26.91  & 33.00  & \textbf{18.00}  & 57.45  & 49.33  & 128.38  & 2369.09 \\
L-206 & BBQ-1 & Lion-CEWT & \textbf{27.21}  & \textbf{33.50}  & 17.49  & \textbf{57.51}  & \textbf{52.88}  & \textbf{113.15}  & \textbf{2148.31} \\ \hline
L-206 & BBQ-1 & SSO      & 26.94  & 33.04  & \textbf{19.03}  & 57.40  & \textbf{52.25}  & 130.55  & 2297.06 \\
L-206 & BBQ-1 & SSO-CEWT & \textbf{27.29}  & \textbf{33.67}  & 17.83  & \textbf{57.40}  & 51.07  & \textbf{122.31}  & \textbf{1961.63} \\ \hline
L-206 & BBQ-2 & MuonH     & 28.71  & 38.13  & 17.06  & \textbf{62.95}  & \textbf{51.14}  & 64.83  & 816.27 \\
L-206 & BBQ-2 & Muon-CEWT & \textbf{28.76}  & \textbf{40.40}  & \textbf{17.83}  & 61.92  & 50.43  & \textbf{63.18}  & \textbf{802.27} \\ \hline
L-206 & BBQ-2 & AdamH     & \textbf{28.78}  & 39.10  & \textbf{19.71}  & \textbf{61.53}  & 50.20  & 66.78  & \textbf{988.03} \\
L-206 & BBQ-2 & Adam-CEWT & 28.72  & \textbf{40.03}  & 18.43  & 60.94  & \textbf{51.93}  & \textbf{63.38}  & 1045.41 \\ \hline
L-206 & BBQ-2 & LionH     & 28.71  & 37.29  & 18.52  & 61.15  & 51.22  & 65.09  & \textbf{1133.95} \\
L-206 & BBQ-2 & Lion-CEWT & \textbf{28.92}  & \textbf{42.00}  & \textbf{19.37}  & \textbf{62.68}  & \textbf{51.22}  & \textbf{62.27}  & 1158.92 \\ \hline
L-206 & BBQ-2 & SSO      & 28.56  & \textbf{40.28}  & \textbf{18.43}  & \textbf{62.24}  & 50.20  & 67.33  & 876.57 \\
L-206 & BBQ-2 & SSO-CEWT & \textbf{28.58}  & 38.05  & 18.17  & 60.94  & \textbf{51.46}  & \textbf{64.47}  & \textbf{799.21} \\ \hline
L-332 & BBQ-1 & AdamH     & 27.33  & 34.60  & \textbf{19.71}  & 58.54  & \textbf{50.43}  & 103.60  & 1512.79 \\
L-332 & BBQ-1 & Adam-CEWT & \textbf{27.44}  & \textbf{35.10}  & 18.43  & \textbf{59.52}  & 50.36  & \textbf{93.12}  & \textbf{1445.00} \\ \hline
L-610 & BBQ-1 & AdamH     & \textbf{28.17}  & 37.50  & \textbf{20.22}  & 59.85  & \textbf{51.22}  & 79.84  & 1041.07 \\
L-610 & BBQ-1 & Adam-CEWT & 27.90  & \textbf{37.67}  & 18.09  & \textbf{60.72}  & 49.57  & \textbf{73.60}  & \textbf{1015.54} \\ \hline
G-204 & BBQ-1 & MuonH     & \textbf{27.03}  & 33.08  & 18.94  & 57.40  & 49.01  & 147.80  & 2428.02 \\
G-204 & BBQ-1 & Muon-CEWT & 26.91  & \textbf{33.33}  & \textbf{19.20}  & \textbf{57.73}  & \textbf{50.83}  & \textbf{134.10}  & \textbf{2020.69} \\ \hline
G-204 & BBQ-1 & AdamH     & \textbf{26.94}  & \textbf{31.94}  & \textbf{18.34}  & 55.88  & 49.57  & 155.48  & 2561.49 \\
G-204 & BBQ-1 & Adam-CEWT & 26.80  & 31.65  & 18.17  & \textbf{57.62}  & \textbf{49.96}  & \textbf{139.19}  & \textbf{2223.63} \\ \hline
G-204 & BBQ-1 & LionH     & 26.77  & 32.53  & 17.58  & \textbf{57.18}  & 50.75  & 149.03  & 2497.89 \\
G-204 & BBQ-1 & Lion-CEWT & \textbf{26.89}  & \textbf{33.04}  & \textbf{18.60}  & 56.09  & \textbf{52.72}  & \textbf{128.73}  & \textbf{1997.89} \\ \hline
G-204 & BBQ-1 & SSO      & \textbf{27.02}  & \textbf{33.50}  & 18.26  & \textbf{57.45}  & \textbf{52.25}  & 141.21  & 2250.85 \\
G-204 & BBQ-1 & SSO-CEWT & 26.90  & 31.94  & \textbf{18.34}  & 57.24  & 50.67  & \textbf{130.86}  & \textbf{1905.77} \\ \hline
G-204 & BBQ-2 & MuonH     & 28.22  & 35.98  & \textbf{19.03}  & 60.17  & 49.25  & 72.07  & 916.26 \\
G-204 & BBQ-2 & Muon-CEWT & \textbf{28.51}  & \textbf{38.43}  & 16.98  & \textbf{60.55}  & \textbf{52.09}  & \textbf{69.59}  & \textbf{822.85} \\ \hline
G-204 & BBQ-2 & AdamH     & 28.01  & 37.84  & 17.32  & \textbf{61.10}  & \textbf{51.85}  & 74.73  & 947.88 \\
G-204 & BBQ-2 & Adam-CEWT & \textbf{28.35}  & \textbf{38.72}  & \textbf{19.62}  & 61.04  & 51.22  & \textbf{71.73}  & \textbf{866.34} \\ \hline
G-204 & BBQ-2 & LionH     & 28.08  & 37.92  & \textbf{19.97}  & 60.39  & 49.96  & 70.75  & 906.43 \\
G-204 & BBQ-2 & Lion-CEWT & \textbf{28.38}  & \textbf{38.26}  & 17.92  & \textbf{61.32}  & \textbf{51.85}  & \textbf{67.67}  & \textbf{831.54} \\ \hline
G-204 & BBQ-2 & SSO      & 28.28  & \textbf{38.43}  & 17.66  & 60.50  & \textbf{52.01}  & 72.15  & 983.45 \\
G-204 & BBQ-2 & SSO-CEWT & \textbf{28.45}  & 38.22  & \textbf{18.69}  & \textbf{61.37}  & 49.88  & \textbf{68.67}  & \textbf{843.55} \\ \hline
G-325 & BBQ-1 & AdamH     & 26.78  & 33.33  & \textbf{19.11}  & \textbf{57.34}  & 50.75  & 123.77  & 1850.39 \\
G-325 & BBQ-1 & Adam-CEWT & \textbf{26.91}  & \textbf{33.54}  & 16.98  & 56.64  & \textbf{51.78}  & \textbf{110.03}  & \textbf{1500.09} \\ \hline
\end{tabular}%
}
\caption{Zero-shot Evaluation of Models (from Table~\ref{tab:main}) Pre-trained on the C4 Dataset.}
\label{tab:zero_shot}
\end{table}

\pagebreak
\section{Training Time}\label{sec:train_time}
Table~\ref{tab:runtime} presents the training time of our experiments in Table~\ref{tab:main}. We note that our experiments were conducted on shared hardware, and the execution time has a non-zero variance. On average (geometric mean), we find that CEWT results in a 4\% longer training time.
\begin{table}[H]
\centering
\resizebox{\textwidth}{!}{%
\renewcommand{\arraystretch}{1.5} 
\begin{tabular}{|l|l|l|cc|cc|cc|cc|}
\hline
M-S & Q-B & GPU & MuonH & Muon-CEWT & AdamH & Adam-CEWT & LionH & Lion-CEWT & SSO & SSO-CEWT \\ \hline
L-95  & L-1 & RTX 5090 & \textbf{2.44} & 2.45  & \textbf{2.42} & 2.43  & \textbf{2.42} & 2.43  & \textbf{3.42} & 3.42 \\ \hline
L-95  & Q-1 & RTX 5090 & \textbf{2.74} & 2.85  & \textbf{2.73} & 2.84  & \textbf{2.72} & 2.83  & \textbf{3.76} & 3.83 \\ \hline
L-95  & B-1 & RTX 5090 & \textbf{2.76} & 2.91  & \textbf{2.74} & 2.90  & \textbf{2.74} & 2.90  & \textbf{3.76} & 3.90 \\ \hline
L-95  & L-2 & RTX 5090 & \textbf{2.44} & 2.46  & \textbf{2.43} & 2.44  & \textbf{2.42} & 2.43  & \textbf{3.42} & 3.42 \\ \hline
L-95  & Q-2 & RTX 5090 & \textbf{2.75} & 2.86  & \textbf{2.74} & 2.84  & \textbf{2.73} & 2.85  & \textbf{3.75} & 3.85 \\ \hline
L-95  & B-2 & RTX 5090 & \textbf{2.77} & 2.92  & \textbf{2.76} & 2.91  & \textbf{2.75} & 2.90  & \textbf{3.75} & 3.90 \\ \hline
L-127 & Q-1 & RTX 5090 & \textbf{6.04} & 6.43  & \textbf{6.00} & 6.38  & \textbf{5.99} & 6.36  & \textbf{8.87} & 9.20 \\ \hline
L-127 & B-1 & RTX 5090 & \textbf{6.06} & 6.55  & \textbf{6.01} & 6.50  & \textbf{6.01} & 6.49  & \textbf{8.84} & 9.31 \\ \hline
L-127 & Q-2 & RTX 5090 & \textbf{6.06} & 6.43  & \textbf{6.02} & 6.38  & \textbf{6.02} & 6.37  & \textbf{8.81} & 9.22 \\ \hline
L-127 & B-2 & RTX 5090 & \textbf{6.09} & 6.57  & \textbf{6.05} & 6.51  & \textbf{6.04} & 6.51  & \textbf{8.82} & 9.31 \\ \hline
L-206 & B-1 & H100 80GB & \textbf{10.28} & 10.76  & \textbf{9.09} & 10.32  & 13.20 & \textbf{10.58}  & \textbf{15.07} & 15.20 \\ \hline
L-206 & B-2 & H100 80GB & 13.15 & \textbf{12.82}  & 13.04 & \textbf{10.68}  & \textbf{10.77} & 14.12  & \textbf{14.99} & 15.36 \\ \hline
L-332 & B-1 & H100 80GB &   -    &   -     & 40.53 & \textbf{38.08}  &   -    &   -     &   -    &   -    \\ \hline
L-610 & B-1 & H100 80GB &   -    &   -     & \textbf{130.52} & 140.41  &   -    &   -     &   -    &   -    \\ \hline
G-94  & L-1 & RTX 5090 & \textbf{2.29} & 2.31  & \textbf{2.28} & 2.29  & \textbf{2.27} & 2.29  & \textbf{3.20} & 3.21 \\ \hline
G-94  & Q-1 & RTX 5090 & \textbf{2.67} & 2.76  & \textbf{2.64} & 2.75  & \textbf{2.64} & 2.74  & \textbf{3.60} & 3.70 \\ \hline
G-94  & B-1 & RTX 5090 & \textbf{2.69} & 2.79  & \textbf{2.66} & 2.77  & \textbf{2.66} & 2.77  & \textbf{3.64} & 4.03 \\ \hline
G-94  & L-2 & RTX 5090 & \textbf{2.30} & 2.31  & \textbf{2.28} & 2.29  & \textbf{2.28} & 2.29  & \textbf{3.22} & 3.22 \\ \hline
G-94  & Q-2 & RTX 5090 & \textbf{2.67} & 2.77  & \textbf{2.65} & 2.75  & \textbf{2.65} & 2.75  & \textbf{3.64} & 3.74 \\ \hline
G-94  & B-2 & RTX 5090 & \textbf{2.69} & 2.80  & \textbf{2.68} & 2.79  & \textbf{2.67} & 2.78  & \textbf{3.65} & 3.76 \\ \hline
G-127 & Q-1 & RTX 5090 & \textbf{5.89} & 6.31  & \textbf{5.83} & 6.26  & \textbf{5.84} & 6.25  & \textbf{8.05} & 8.45 \\ \hline
G-127 & B-1 & RTX 5090 & \textbf{5.94} & 6.26  & \textbf{5.88} & 6.22  & \textbf{5.87} & 6.21  & \textbf{8.09} & 8.42 \\ \hline
G-127 & Q-2 & RTX 5090 & \textbf{5.91} & 6.32  & \textbf{5.86} & 6.27  & \textbf{5.86} & 6.26  & \textbf{8.06} & 8.50 \\ \hline
G-127 & B-2 & RTX 5090 & \textbf{5.96} & 6.29  & \textbf{5.91} & 6.25  & \textbf{5.91} & 6.23  & \textbf{8.08} & 8.43 \\ \hline
G-204 & B-1 & H100 80GB & 10.43 & \textbf{9.73}  & \textbf{9.98} & 10.43  & 9.66 & \textbf{9.25}  & \textbf{13.59} & 19.75 \\ \hline
G-204 & B-2 & H100 80GB & \textbf{8.77} & 10.40  & \textbf{8.76} & 8.93  & \textbf{9.52} & 10.05  & \textbf{13.73} & 18.66 \\ \hline
G-325 & B-1 & H100 80GB &   -    &   -     & 38.12 & \textbf{35.41}  &   -    &   -     &   -    &   -    \\ \hline
\end{tabular}%
}
\caption{Training Time in Hours of experiments in Table~\ref{tab:main}.}
\label{tab:runtime}
\end{table}

\section{Evaluation on Vision Models}\label{sec:vision}
While our main empirical claims are on LLMs, we additional evaluate CEWT with the vision model DEIT-Base~\citep{pmlr-v139-touvron21a} pre-trained on Imagenet-64x64~\citep{russakovsky2015imagenetlargescalevisual,chrabaszcz2017downsampledvariantimagenetalternative}. Following \citet{pmlr-v139-touvron21a}, we use 16x16 pixels as a token, and therefore each 64x64 image is converted to a flattened sequence of 16 tokens. We use a batch size of 1024, a hypersphere learning rate of 0.004, and AdamH/Adam-CEWT to conduct our experiments. 

\begin{table}[!htbp]
\centering
\resizebox{\textwidth}{!}{%
\renewcommand{\arraystretch}{1.2} 
\begin{tabular}{|l|l|l|l|l|l|c|c|c|}
\hline
Exp ID & Model     & Epochs & Quantizer & A/W & Optimizer & RBM & Train Loss    & Test Accuracy  \\ \hline
1 & DEIT-Base & 300 & BBQ   & 16/2 & AdamH     &        {0.018} &        {4.25} &        {60.22} \\
2 & DEIT-Base & 300 & BBQ   & 16/2 & Adam-CEWT & \textbf{0.010} & \textbf{4.15} & \textbf{61.04} \\ \hline
3 & DEIT-Base & 300 & BBQ   & 16/1 & AdamH     &        {0.020} &        {4.54} &        {56.41} \\
4 & DEIT-Base & 300 & BBQ   & 16/1 & Adam-CEWT & \textbf{0.010} & \textbf{4.39} & \textbf{58.88} \\ \hline
5 & DEIT-Base & 300 & QuEST & 16/2 & AdamH     &        {0.088} &        {4.19} & \textbf{61.32} \\
6 & DEIT-Base & 300 & QuEST & 16/2 & Adam-CEWT & \textbf{0.056} & \textbf{4.16} &        {61.18} \\ \hline
7 & DEIT-Base & 300 & QuEST & 16/1 & AdamH     &        {0.213} &        {4.55} &        {57.08} \\
8 & DEIT-Base & 300 & QuEST & 16/1 & Adam-CEWT & \textbf{0.119} & \textbf{4.41} & \textbf{58.70} \\ \hline \hline
9 & DEIT-Base & 300 & BBQ   & 2/2 & AdamH     &        {0.011} &        {4.56} &        {55.85} \\
10 & DEIT-Base & 300 & BBQ   & 2/2 & Adam-CEWT & \textbf{0.010} & \textbf{4.53} & \textbf{56.60} \\ \hline
11 & DEIT-Base & 300 & BBQ   & 1/1 & AdamH     &        {0.013} & \textbf{5.34} & \textbf{39.80} \\
12 & DEIT-Base & 300 & BBQ   & 1/1 & Adam-CEWT & \textbf{0.010} & \textbf{5.34} &        {39.24} \\ \hline
13 & DEIT-Base & 300 & QuEST & 2/2 & AdamH     &        {0.081} & \textbf{4.57} & \textbf{56.51} \\
14 & DEIT-Base & 300 & QuEST & 2/2 & Adam-CEWT & \textbf{0.056} &        {4.58} &        {56.06} \\ \hline
15 & DEIT-Base & 300 & QuEST & 1/1 & AdamH     &        {0.215} &        {5.37} &        {38.71} \\
16 & DEIT-Base & 300 & QuEST & 1/1 & Adam-CEWT & \textbf{0.119} & \textbf{5.35} & \textbf{39.39} \\ \hline\hline
17 & DEIT-Base & 600 & BBQ   & 2/2 & AdamH     &        {0.014} & \textbf{4.38} & \textbf{58.19} \\
18 & DEIT-Base & 600 & BBQ   & 2/2 & Adam-CEWT & \textbf{0.010} & \textbf{4.38} &        {58.16} \\ \hline
19 & DEIT-Base & 600 & BBQ   & 1/1 & AdamH     &        {0.017} &        {5.25} &        {42.10} \\
20 & DEIT-Base & 600 & BBQ   & 1/1 & Adam-CEWT & \textbf{0.010} & \textbf{5.20} & \textbf{42.92} \\ \hline
21 & DEIT-Base & 600 & QuEST & 2/2 & AdamH     &        {0.084} &        {4.44} &        {58.21} \\
22 & DEIT-Base & 600 & QuEST & 2/2 & Adam-CEWT & \textbf{0.056} & \textbf{4.41} & \textbf{58.49} \\ \hline
23 & DEIT-Base & 600 & QuEST & 1/1 & AdamH     &        {0.219} &        {5.29} &        {40.54} \\
24 & DEIT-Base & 600 & QuEST & 1/1 & Adam-CEWT & \textbf{0.119} & \textbf{5.26} & \textbf{41.53} \\ \hline
\end{tabular}%
}
\caption{Evaluation on Vision Models pre-trained on Imagenet 64x64.}
\label{tab:vit}
\end{table}

We first run weight-only quantization experiments with IDs 1-8 in Table~\ref{tab:vit}. Results confirm that rounding boundary weight oscillation can decrease training efficiency, and that CEWT can suppress weight oscillation and achieve lower loss. 

We next run weight-activation quantization experiments with IDs 9-16 in Table~\ref{tab:vit}. Results suggest that rounding boundary weight oscillation is less severe (when CEWT is not used) and therefore the loss reduction of CEWT is less consistent. We believe weight oscillation is less severe because activation quantization can reduce training efficiency (especially in a training recipe with heavy data augmentation) and delay the behaviour of weight oscillation to a much later stage in training. Table~\ref{tab:vit_rbm} shows that the rounding boundary mass consistently increases (throughout training) in weight-activation quantization experiments, but the rate of RBM increase is much slower compared to weight-only quantization experiments. Table~\ref{tab:vit_rbm} suggests that 300 epochs are not enough to train activation-weight-quantized DEIT-Base models to full convergence.

\begin{table}[!htbp]
\centering
\resizebox{\textwidth}{!}{%
\renewcommand{\arraystretch}{1.2} 
\begin{tabular}{|l|l|l|l|l|l|l|l|l|l|l|}
\hline
Exp ID & Model     & Quantizer & A/W & Optimizer & Ep. 50 & Ep. 100    & Ep. 150 & Ep. 200 & Ep. 250 & Ep. 299  \\ \hline
1 & DEIT-Base & BBQ   & 16/2 & AdamH     &        {0.008} &        {0.010} &        {0.012} &        {0.013} &        {0.012} &        {0.018} \\
9 & DEIT-Base & BBQ   &  2/2 & AdamH     & \textbf{0.006} & \textbf{0.006} & \textbf{0.006} & \textbf{0.006} & \textbf{0.007} & \textbf{0.011} \\ \hline
2 & DEIT-Base & BBQ   & 16/1 & AdamH     &        {0.005} &        {0.008} &        {0.009} &        {0.009} &        {0.010} &        {0.020} \\
11 & DEIT-Base & BBQ   &  1/1 & AdamH     & \textbf{0.003} & \textbf{0.003} & \textbf{0.003} & \textbf{0.004} & \textbf{0.005} & \textbf{0.013} \\ \hline
3 & DEIT-Base & QuEST & 16/2 & AdamH     &        {0.077} &        {0.079} &        {0.079} &        {0.079} &        {0.079} &        {0.088} \\
13 & DEIT-Base & QuEST &  2/2 & AdamH     & \textbf{0.067} & \textbf{0.067} & \textbf{0.068} & \textbf{0.069} & \textbf{0.071} & \textbf{0.081} \\ \hline
4 & DEIT-Base & QuEST & 16/1 & AdamH     &        {0.193} &        {0.198} &        {0.200} &        {0.201} &        {0.203} & \textbf{0.213} \\
15 & DEIT-Base & QuEST &  1/1 & AdamH     & \textbf{0.186} & \textbf{0.187} & \textbf{0.189} & \textbf{0.192} & \textbf{0.198} &        {0.215} \\ \hline

\end{tabular}%
}
\caption{RBM of Vision Models at different epochs (Ep.). The experiment IDs in this table refers to the IDs from Table~\ref{tab:vit}. }
\label{tab:vit_rbm}
\end{table}

Therefore, we double the number of epochs (from 300 to 600) and run weight-activation quantization experiments with IDs 17-24 in Table~\ref{tab:vit}. Results suggest that rounding boundary weight oscillation is more severe (when CEWT is not used) compared to experiments with 300 epochs, and CEWT can now consistently achieve lower loss, confirming our insight from Section~\ref{sec:sweep_dniter} and Table~\ref{tab:toy_sweep_iter}.

Overall, our results suggest that rounding boundary weight oscillation becomes more severe as training time increases, and therefore the loss reduction effect of CEWT becomes more consistent with longer training.

\clearpage
\endgroup

\end{document}